%% file: example_paper.tex
\documentclass{article}

\usepackage{microtype}
\usepackage{graphicx}
\usepackage{subfig}
\usepackage{booktabs} 
\usepackage{tablefootnote}
\usepackage{hyperref}
\usepackage{multirow}

\usepackage[accepted]{icml2024}

\usepackage{amsmath}
\usepackage{amssymb}
\usepackage{mathtools}
\usepackage{amsthm}

\usepackage[capitalize,noabbrev]{cleveref}

\theoremstyle{plain}

\theoremstyle{definition}

\theoremstyle{remark}

\usepackage[textsize=tiny]{todonotes}

\icmltitlerunning{Fast Text-to-3D-Aware Face Generation and Manipulation via Direct Cross-modal Mapping and Geometric Regularization}

\begin{document}

\twocolumn[
\icmltitle{Fast Text-to-3D-Aware Face Generation and Manipulation\\ via Direct Cross-modal Mapping and Geometric Regularization}



\icmlsetsymbol{equal}{*}

\begin{icmlauthorlist}
\icmlauthor{Jinlu Zhang}{equal,auth1}
\icmlauthor{Yiyi Zhou}{equal,auth1}
\icmlauthor{Qiancheng Zheng}{auth1}
\icmlauthor{Xiaoxiong Du}{auth1}
\\
\icmlauthor{Gen Luo}{auth1}
\icmlauthor{Jun Peng}{auth2}
\icmlauthor{Xiaoshuai Sun}{auth1}
\icmlauthor{Rongrong Ji}{auth1}
\end{icmlauthorlist}

\icmlaffiliation{auth1}{Key Laboratory of Multimedia Trusted Perception and Efficient Computing, Ministry of Education of China, Xiamen University, 361005, P.R. China.}
\icmlaffiliation{auth2}{Peng Cheng Laboratory, Shenzhen, 518000, P.R. China}

\icmlcorrespondingauthor{Xiaoshuai Sun}{xssun@xmu.edu.cn}

\icmlkeywords{Generative Model, Cross-Modal Mapping, controllable generation}

\vskip 0.3in
]



\printAffiliationsAndNotice{\icmlEqualContribution} 

\input{sec/0_abs}  
\input{sec/1_intro}
\input{sec/2_rela}

\input{sec/3_method}

\input{sec/4_exp}
\input{sec/5_conclusion}

\section*{Acknowledgements}
This work was supported by National Key R\&D Program of China (No.2022ZD0118201), the National Science Fund for Distinguished Young Scholars (No.62025603), the National Natural Science Foundation of China (No. U21B2037, No. U22B2051, No.U21A20472, No. 62072389,  No. 623B2088, No. 62176222, No. 62176223, No. 62176226, No. 62072386, No. 62072387, No. 62002305 and No. 62272401), the National Natural Science Fund for Young Scholars of China (No. 62302411), China Postdoctoral Science Foundation (No. 2023M732948), the Natural Science Foundation of Fujian Province of China (No.2021J01002,  No.2022J06001), and partially sponsored by CCF-NetEase ThunderFire Innovation Research Funding (NO. CCF-Netease 202301).


\section*{Impact Statement}
This paper enhances the visual quality and improve the efficiency of text-to-3D-aware face generation through \emph{End-to-End cross modal mapping}, which will further advance the development of 3D-aware image generation. Our work may have various potential societal consequences, though none we believe need to be specifically highlighted here.

\nocite{langley00}

\bibliography{example_paper}
\bibliographystyle{icml2024}

\newpage
\appendix
\onecolumn
\input{sec/x_appendix}

\end{document}

%% file: sec/0_abs.tex
\begin{abstract}
Text-to-3D-aware face (T3D Face) generation and manipulation is an emerging research hot spot in machine learning, which still suffers from low efficiency and poor quality. In this paper, we propose an \emph{\textbf{E}nd-to-End \textbf{E}fficient and \textbf{E}ffective} network for fast and accurate T3D face generation and manipulation, termed $E^3$-FaceNet. Different from existing complex generation paradigms, $E^3$-FaceNet resorts to a direct mapping from text instructions to 3D-aware visual space. We introduce a novel \emph{Style Code Enhancer} to enhance cross-modal semantic alignment, alongside an innovative \emph{Geometric Regularization} objective to maintain consistency across multi-view generations. Extensive experiments on three benchmark datasets demonstrate that $E^3$-FaceNet can not only achieve picture-like 3D face generation and manipulation, but also improve inference speed by orders of magnitudes. For instance, compared with Latent3D, $E^3$-FaceNet speeds up the five-view generations by almost 470 times, while still exceeding in generation quality. Our code is released at \url{https://github.com/Aria-Zhangjl/E3-FaceNet}.
\end{abstract}

%% file: sec/1_intro.tex
\begin{figure*}[!h]
    \includegraphics[width=1.\textwidth]{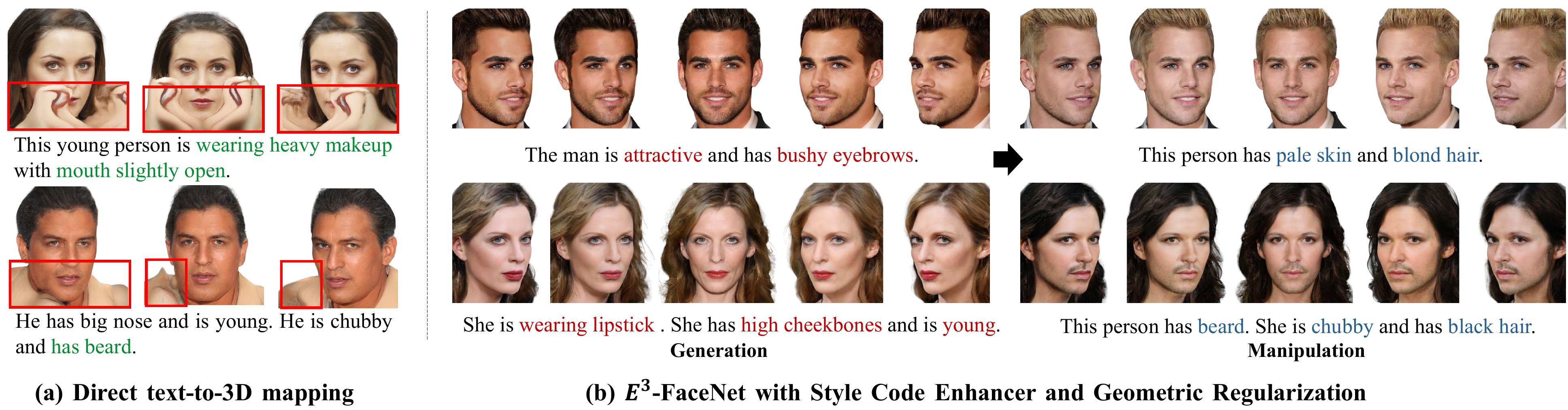}
    \vspace{-6.5mm}
    \caption{\textbf{(a)} The 3D-aware portraits generated by direct cross-modal mapping. Without any 3D regularization and semantic enhancement, the model tends to generate faces with artifacts (labeled in \textcolor[RGB]{192,0,0}{\textbf{red}} box) and fails to generate the accurate face attribute (shown in \textcolor[RGB]{1,141,48}{\textbf{green}} color). \textbf{(b)} Multi-view results of T3D face generation and manipulation by our $E^3$-FaceNet. With the proposed Style Code Enhancer and Geometric Regularization, $E^3$-FaceNet can well align the text instructions and generate high-quality 3D-aware face images without obvious defects (\textcolor[RGB]{192,0,0}{\textbf{red}} for generation and \textcolor[RGB]{38,95,146}{\textbf{blue}} for manipulation). Notably, the speeds of five-view image generation and manipulation of $E^3$-FaceNet are only 0.64s and 0.89s, respectively. The backgrounds of these images are removed for better comparison.}
    \label{fig:FIG1}
    \vspace{-3.5mm}
\end{figure*}

\section{Introduction}
\label{sec:intro}
Text-to-3D-aware face (T3D Face) generation is an emerging research hot spot in machine learning \cite{toshpulatov2021generative,zhang2023dreamface,wu2023high,xia2023survey}. Compared with the well-developed 2D face generation \cite{xia2021tedigan,patashnik2021styleclip,peng2022learning,peng2022towards,du2023pixelface+}, T3D Face not only requires to depict the intricate details of different facial components but also to model the 3D shapes and positions under different poses \cite{toshpulatov2021generative,xia2023survey}.  Additionally, the semantic consistency between cross-modal \cite{zhou2019plenty, luo2022towards, zhou2021real, wu2023high} and cross-view \cite{ or2022stylesdf, chan2022efficient,xia2023survey} generations is also a main obstacle.

To overcome these challenges, most of existing methods often adopt a multi-stage pipeline for high-quality T3D Face generation \cite{zhang2023dreamface,wu2023high,rowan2023text2face,li20233d}. For instance, \citet{wu2023high} propose two independent networks to generate face polygon mesh and texture map separately, and then synthesize the textured mesh by the renderer \cite{li2018differentiable}. \citet{li20233d} first generate the latent code via a diffusion model \cite{rombach2022high} and then feed it to a pre-trained 3D generator for T3D Face generations. Moreover, most methods also require test-time tuning in 2D image space \cite{zhang2023dreamface,canfes2023text,zhang2023fdnerf} or the manual editing of facial attributes \cite{wu2023high}. Despite effectiveness, the training and inference of existing methods are still time-consuming, greatly impeding their real-world applications.

Unlike previous works, we are keen to fast and accurate T3D face generation and manipulation via direct cross-modal mapping. In particular, we aim to use the text description to impact the whole process of 3D face generation, including the learning of the implicit 3D scene representations, \emph{i.e.}, NeRF \cite{mildenhall2021nerf}, and the subsequent 2D upsampling. This intuition has been well validated in the recent progress of T2D face generation \cite{peng2022learning,sun2022anyface,peng2022towards}, where the text feature can be directly used to modulate the sampled latent code for adversarial image generation. Adopting this direct cross-modal mapping strategy can help T3D Face methods avoid the complex processing steps in existing multi-stage pipelines, thereby substantially improving inference speed.

However, direct text-to-3D modeling is still intractable, which suffers from two main issues. First, most T3D face datasets contain only single-view face images aligned with text descriptions, lacking accurately matched T3D face examples from multi-view collections \cite{yu2023towards}. Consequently, a cross-modal mapping T3D face model is often hard to be fully supervised during training, leading to the generation of obvious artifacts, as shown in Fig.\ref{fig:FIG1}(a). In addition, the visual semantic space of T3D face generation is much larger than that of T2D face, so it is more challenging to ensure semantic consistency among cross-modal and cross-view generations. For instance, the model is prone to overlooking the attributes in the text prompt, as shown in Fig.\ref{fig:FIG1}(a). In this case, direct mapping for T3D face still remains an open problem. 

To remedy these issues, we propose a novel \emph{End-to-end Efficient and Effective} T3D Face generation and manipulation network in this paper, termed $E^3$-FaceNet. Concretely, $E^3$-FaceNet is built based on a pre-trained unconditional 3D generation network called \emph{StyleNeRF} \cite{gu2021stylenerf}. In $E^3$-FaceNet, the text representation is directly applied to modulate the sampled noise for end-to-end text-guided 3D generation. To improve semantic alignment, we also propose a novel \emph{Style Code Enhancer} to inject text information into the process of adversarial image rendering, which also enables $E^3$-FaceNet with the ability of text-driven face editing. Meanwhile, to generate high-quality 3D geometric details, we also propose a novel learning objective for $E^3$-FaceNet, which leverages both basic and high-order geometric information to regularize the generated faces in 3D space, thereby helping the model synthesize more vibrant and natural-looking 3D faces. Equipped with these two innovative designs, $E^3$-FaceNet achieves efficient and precise controls over 3D face synthesis, as shown in Fig.\ref{fig:FIG1}(b). 

To validate our $E^3$-FaceNet, we conduct extensive experiments on three widely-used benchmarks, namely \emph{Multi-Modal CelebA-HQ} \cite{xia2021tedigan}, \emph{CelebAText-HQ} \cite{sun2021multi} and \emph{FFHQ-Text} \cite{zhou2021generative}, and compare $E^3$-FaceNet with a set of state-of-the-art (SOTA) methods of 2D and 3D face generations, including \cite{xia2021tedigan,peng2022learning,peng2022towards,wu2023high,aneja2023clipface}. The experimental results show that our method not only performs better than existing T3D face methods in generation quality, 3D consistency and semantic alignment, but also improves the inference speed by orders of magnitude \emph{e.g.}, 0.64\emph{s} \emph{v.s.} 302\emph{s} of Latent3D \cite{canfes2023text}. Compared with T2D face methods, $E^3$-FaceNet also excels in single-view image quality, \emph{e.g.}, 12.46 FID \emph{v.s.} 53.38 FID of StyleCLIP \cite{patashnik2021styleclip} on MMCelebA-HQ. Moreover, our method also supports accurate 3D face manipulation with very limited time cost, \emph{e.g}, only 0.17\emph{s} per view. These results well validate the effectiveness and efficiency of $E^3$-FaceNet towards fast and accurate T3D Face generation and manipulation.

Conclusively, the contribution of this paper is three-fold:
\begin{itemize}
    \item We propose a novel cross-modal mapping method for fast and accurate T3D-Face generation and manipulation, termed $E^3$-FaceNet. 
    \item We propose a novel \emph{geometric regularization} to avoid the artifacts caused by direct cross-modal mapping and an innovative module called \emph{Style Code Enhancer} to ensure the semantic consistency of T3D generation. 
    \item  On a bunch of 2D and 3D benchmarks, our $E^3$-FaceNet outperforms a set of compared methods in terms of image quality and semantic consistency, and improves the speed of T3D Face generation by orders of magnitude.
\end{itemize}

%% file: sec/2_rela.tex
\section{Related Work}
\label{sec:rw}
\subsection{3D-aware Image Generation}
The task of 3D-aware image generation \cite{xia2023survey} is to generate high-quality renderings that are consistent across multiple views. 
Recent studies have been conducted in representing 3D scenes with neural implicit functions\cite{mescheder2019occupancy,park2019deepsdf,mildenhall2021nerf}, among which \emph{neural implicit representation} (NIR) \cite{mescheder2019occupancy,mildenhall2021nerf,park2019deepsdf} emerges as an effective alternative for learning-based 3D reconstruction with multi-view 2D supervision. This representation is then applied to 3D image synthesis under single-view image supervision \cite{schwarz2020graf, chan2021pi, gu2021stylenerf, or2022stylesdf, chan2022efficient}. In this paper, we extend the representative implicit method, \emph{i.e.,} StyleNeRF \cite{gu2021stylenerf}, for text-conditioned 3D face generation. By incorporating text semantics into the latent code, our proposed method enables the end-to-end mapping from text words to 3D faces.
 \begin{figure*}[t]
    \centering
    \includegraphics[width=1.0\textwidth]{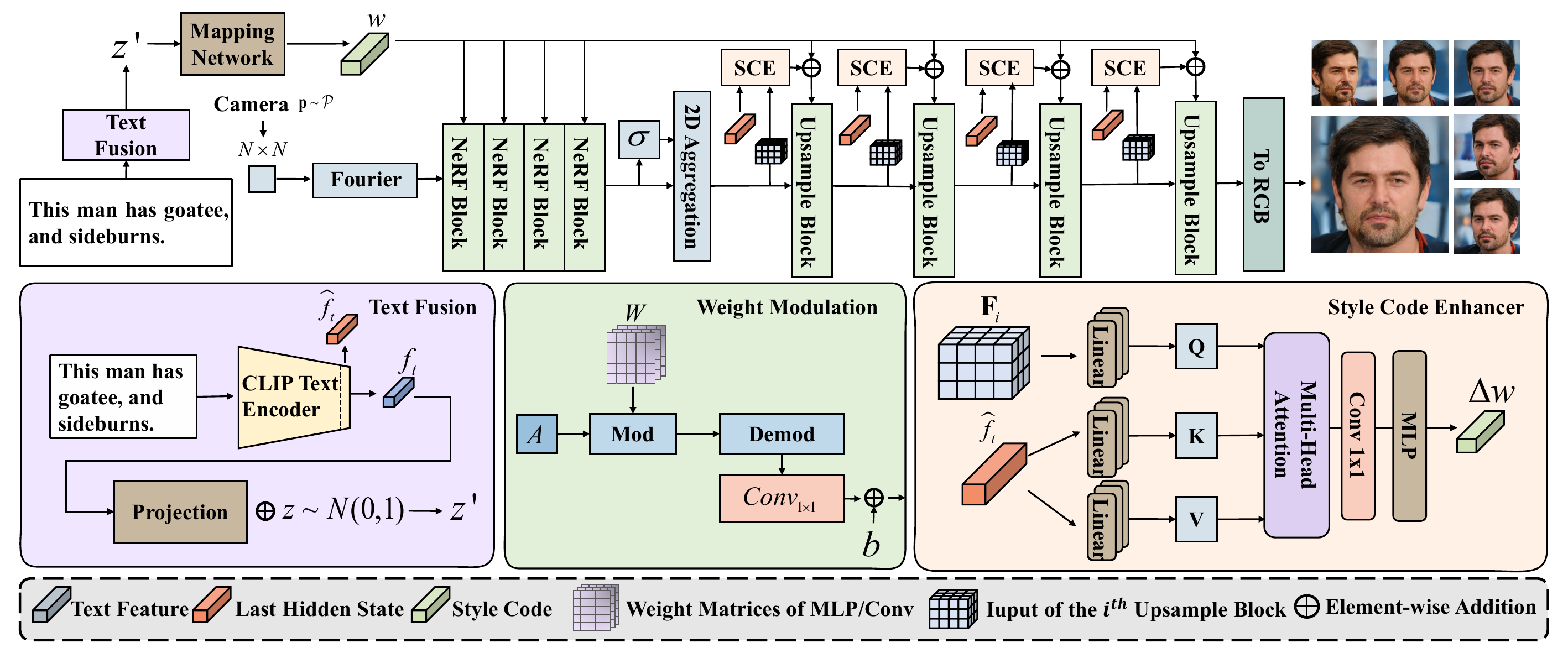}
    \vspace{-6mm}
    \caption{The overall structure of $E^3$-FaceNet. $E^3$-FaceNet uses the extracted text features to module the sampled noise, thereby achieving text-guided 3D-aware generation. To improve fine-grained semantic alignment, we introduce a \emph{Style Code Enhancer} (SCE) to predict style code offsets in each upsample block, which also enables $E^3$-FaceNet to perform fast face editing through the interpolation of offsets. }
    \label{fig:architecture}
\vspace{-2.mm}
\end{figure*}
\subsection{Text-to-3D Face Generation}
Text-to-3D face (T3D Face) generation often aims to incorporate text information into 3D generative models for language-guided 3D face generation. Dreamface \cite{zhang2023dreamface} and Describe3D \cite{wu2023high} propose to generate text-conditioned texture maps to render 3D morphable models (3DMM) \cite{blanz2023morphable}.  Despite the effectiveness of geometric modeling, these methods require additional processes to add more facial assets like hair or wrinkle. A relevant work to this paper is TG-3DFace \cite{yu2023towards}, which uses tri-plane neural representations and extends EG3D \cite{chan2022efficient}, a 3D-aware GAN, for end-to-end text-conditioned generation. However, its text-control mechanism is still compositional, which needs to additionally integrate a face parser and train an attribute classifier for fine-grained semantics injection. To guide the generator in learning correct priors, TG-3DFace incorporates a camera pose conditional discriminator during the training process. Different from previous methods, our proposed $E^3$-FaceNet seamlessly integrates text information into the whole process of 3D representation and 2D upsampling, achieving single-stage cross-modal mapping without any further interventions. Besides, we also propose a set of innovative designs to cope with the problems encountered in direct T3D Face generations, \emph{i.e.}, 3D geometric regularization and Style Code Enhancer. 
\subsection{Text-Guided 3D Face Manipulation}
Despite the advancement \cite{jetchev2021clipmatrix,michel2022text2mesh,aneja2023clipface,canfes2023text}, text-guided 3D face manipulation still have obvious drawbacks in inference speed and generalization. Build upon TB-GAN \cite{gecer2020synthesizing}, Latent3D \cite{canfes2023text} optimizes an intermediate layer using a CLIP-based loss \cite{radford2021learning} to generate UV-texture maps. However, the editing quality for textures and expressions are limited due to reliance on 3D scan data. Another method, \emph{i.e.,} ClipFace \cite{aneja2023clipface}, uses two mappers to predict texture and expression latent codes but requires training new mappers for each text instruction, greatly limiting its efficiency and generalizations. In contrast, our $E^3$-FaceNet can handle different editing instructions via  predicting the style code offset without instance level optimization, achieving efficient and effective 3D face manipulation.

%% file: sec/3_method.tex
\section{Method}
\label{sec:method}
\subsection{Preliminary}
\label{subsec:NeRF}
We first recap the preliminaries of our $E^3$-FaceNet, including \emph{Neural Radiance Field} (NeRF) \cite{mildenhall2021nerf} and the 3D generation network \emph{StyleNeRF} \cite{gu2021stylenerf}.

NeRF represents a continuous scene as function $\mathcal{F}(\mathbf{x},\mathbf{v})=({c},\sigma)$, which takes a 3D point $\mathbf{x}\in \mathbf{R}^3$ and a viewing direction $\mathbf{v}\in\mathbf{R}^2$ as input and a RGB color $c\in[0,1]^3$ together with a scalar volume density $\sigma \in[0,\infty]$ as output. To better model high-frequency details, \citet{mildenhall2021nerf} map each dimension of $\mathbf{x}$ and $\mathbf{v}$ to high-dimension feature space with Fourier features $\zeta$. 
For given near and far bounds $t_n$ and $t_f$, the expected rendered color $\hat{\mathbf{c}}_{\theta}$ of ray $\mathbf{r}(t)=\mathbf{o}+t\mathbf{v}$ is obtained by
\begin{equation}
\hat{\mathbf{c}}_{\theta}(\mathbf{r})=\int_{t_n}^{t_f}T(t)\sigma(\mathbf{r}(t))\mathbf{c}(\mathbf{r}(t),\mathbf{v})dt,
\label{equ:render_color}
\end{equation}
where
\begin{equation}
    T(t)=\exp\!\left(-\int_{t_n}^t\sigma(\mathbf{r}(s))ds\right).
\end{equation}

In practice, NeRF formulates $\mathcal{F}$ as MLPs and the parameters $\theta$ is optimized over a set of input images and their camera poses by minimizing
\begin{equation}
    L =\sum_{\mathbf{r}\in\mathcal{R}_{i}}\|\hat{\mathbf{c}}_{\theta}(\mathbf{r})-\mathbf{c}_{\mathrm{GT}}(\mathbf{r})\|^{2}
\end{equation}
where $\mathcal{R}_{i}$ denotes a set of input rays and $\mathbf{c}_{\mathrm{GT}}$ is the corresponding GT color. 

To enable the control of style attributes, StyleNeRF formalizes 3D representations by conditioning NeRF with a style code $w=f(z), z \in \mathcal{N}(0,I)$ as
\begin{equation}
\begin{aligned}
\phi_{\boldsymbol{w}}^n(x)=g_{\boldsymbol{w}}^n\circ \ldots\circ g_{\boldsymbol{w}}^1\circ\zeta\left(\mathbf{x}\right),
\label{equ:stylenerf-representation}
\end{aligned}
\end{equation}
where $\phi_{\boldsymbol{w}}^{n}(\mathbf{x})$ is the $n^{th}$ layer feature of point $\mathbf{x}$, $f$ is the mapping network that projects the noise vector $z$ to the style space $\mathcal{W}$, and $g_w^i(.)$ is the $i^{th}$ MLP modulated by $w$. 

Then the final predicted RBG value $c$ and density $\sigma$ can be obtained by
\begin{equation}
\resizebox{.9\hsize}{!}{$\sigma_{\boldsymbol{w}}(\mathbf{x})=h_{\sigma}\circ\phi_{\boldsymbol{w}}^{n_{\sigma}}(\mathbf{x}), c_{\boldsymbol{w}}(\mathbf{x},\boldsymbol{v})=h_{c}\circ\left[\phi_{\boldsymbol{w}}^{n_{c}}(\mathbf{x}),\zeta\left(\boldsymbol{v}\right)\right]$},
\end{equation}
where $h_{\sigma}$ and $h_{c}$ are linear projections. In order to improve efficiency, StyleNeRF first renders a low-resolution feature map by approximating Eq.\ref{equ:render_color} as:
\begin{equation}
    \hat{\mathbf{c}}_{\theta}(\mathbf{r})=h_c\circ\left[\phi_{\boldsymbol{w}}^{n_c,n_\sigma}\left(\mathcal{A}(r)\right),\zeta\left(\mathbf{v}\right)\right],
\end{equation}
where $\phi_{\boldsymbol{w}}^{n,n_{\sigma}}\left(\mathcal{A}(\boldsymbol{r})\right)=g_{\boldsymbol{w}}^{n}\circ \ldots\circ g_{\boldsymbol{w}}^{n_{\sigma}+1}\circ\mathcal{A}(\boldsymbol{r})$ and $\mathcal{A}({r})=\int_{t_n}^{t_f}T(t)\cdot\sigma_{w}(r(t))\cdot\phi_{w}^{n_{\sigma}}({r}(t))dt$. 
These aggregated features will be upsampled to the desired resolution in 2D space. 

By incorporating NeRF into the style-based architecture, StyleNeRF supports style mixing and simple editing by exploring the style space, indicating that this latent space is semantically editable. Inspired by StyleNeRF and previous adversarial T2D face methods \cite{sun2022anyface,peng2022towards}, we resort to mapping text information onto this style space for precise language-guided 3D generation.

\subsection{The Overall Framework}
\label{subsec:framework}
The overall structure of $E^3$-FaceNet is depicted in Fig.\ref{fig:architecture}. In principle, $E^3$-FaceNet aims to build a deterministic function to directly map the text description $T$ and the noise vector $z$ to the 3D-aware face image $I$ of the camera pose $p$: 
\begin{equation}
    G: (z,T,p) \to I,
\label{equ:mapping}
\end{equation}
where $G$ denotes the generation network, $z\in Z$ is sampled from \emph{Gaussian} latent space, and $p \in P$ is the camera pose.

To realize Eq.\ref{equ:mapping}, we first extract the feature of the description $f_{t}$ by the CLIP text encoder and map it to the latent space $Z$. Then we modulate the random noise $z \in \mathcal{R}^{d}$ by $z' = z \oplus f_t'$, where $f_t'$ is the projected text feature and $\oplus$ denote to element-wise addition. 

Subsequently, the corresponding style code $w$ is obtained by a mapping network $w=f(z') \in \mathcal{W}$. In this simple way, we can effectively combine text information with the style code, and impact both 3D and 2D representation learnings. 

However, this injection is still of limited influence, and achieving well alignment between the synthesis and text semantics remains a challenge  \cite{yu2023towards,wu2023high}. Meanwhile, an inherent problem of training a 3D generative model using 2d unposed supervision is the absence of multi-view information \cite{gu2021stylenerf,xia2023survey}. As shown in Fig.\ref{fig:FIG1}(a), the model tends to generate 3D faces with obvious artifacts and low semantic alignment. To overcome these challenges, we further propose a novel \emph{Style Code Enhancer} and a \emph{geometric regularization}. 

\subsection{Style Code Enhancer}
\label{subsec:stylecode_enhancer}
 As discussed above, it is often difficult to control the entire generation with a single style code $w$ in 2D image generation \cite{abdal2020image2stylegan++,saha2021loho}. And this problem will become more prominent in T3D face generation, since the 3D visual semantic space is much larger than that of the 2D one \cite{xia2023survey}. Previous studies \cite{karras2019style, patashnik2021styleclip, xia2022gan} have shown the varying levels of detail captured by different layers in a 2D image generator.   Based on this, we divide the T3D face generation process into two parts: NeRF blocks generate coarse textual semantics like facial contour and shape, and 2D upsample blocks render intricate facial attributes. We introduce a \emph{Style Code Enhancer} (SCE) to enhance fine-grained text semantics incorporation and improve the impact of style codes in the rendering process. 
 

Concretely, SCE at the $i^{th}$ upsample block first fuses the input image features $\mathbf{F}_{i}$ with the text one $\hat{f}_t$ via a cross-modal attention \cite{rombach2022high}:
\begin{equation} 
    \begin{aligned}
        \mathbf{F'}_{i} &= Attention(Q^i,K^i,V^i) \\&= softmax(\frac{Q^i{K^i}^T}{\sqrt{d}}\cdot V^i),
    \end{aligned}
    \label{equ:sce_ca1}
\end{equation}
with
\begin{gather}
    Q^i=W^i_Q \cdot \mathbf{F}_{i}, K^i=W^{i}_K \cdot \hat{f}_t, V^i = W^{i}_V \cdot \hat{f}_t.
    \label{equ:sce_ca2}
\end{gather}
Here, $\hat{f}_t$ refers to the last hidden state of the text encoder. 

\begin{figure}[t]
    \centering
    \includegraphics[width=0.45\textwidth]{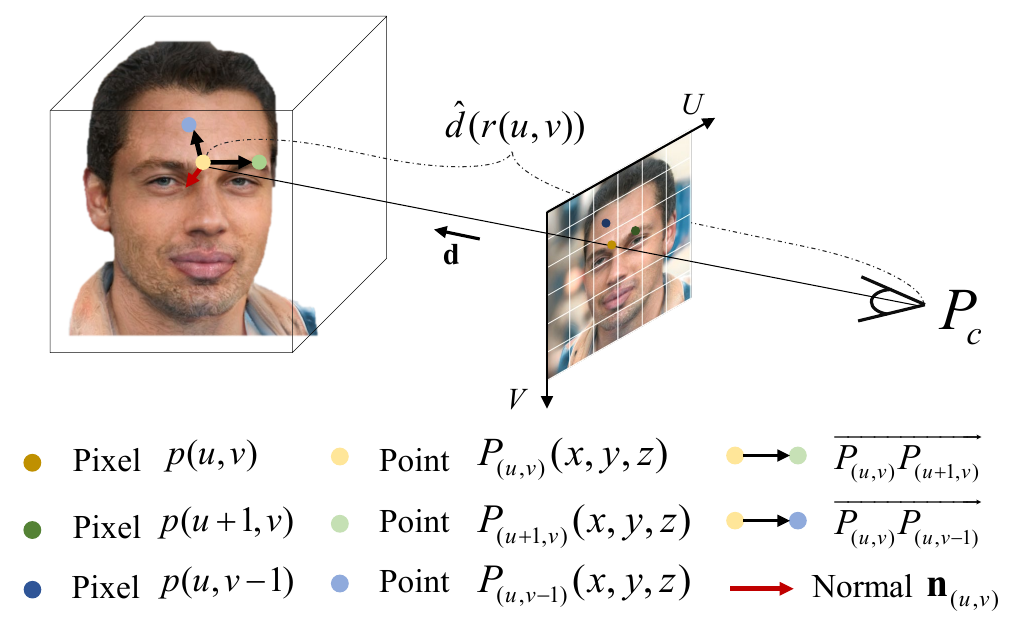}
    \caption{Illustration of the corresponding 3D location $P_{(u,v)}$ and normal vector $n_{(u,v)}$ for a given pixel $p_{(u,v)}$. To generate smooth and natural-looking faces with the absence of multi-view information, we propose to regularize these two terms in 3D space.}
    \label{fig:geometric}
\vspace{-2mm}
\end{figure}

Then $\mathbf{F'}_{i}$ is used to predict the offset $\Delta w_{i}$ of the style code, denoted as $w_{i}' = w + \Delta w_{i}$, as well as to modulate the current upsample block by 
\begin{equation}
    w_{i}' = w+ \Delta w_i = w + MLP(Conv_{1\times 1}(\mathbf{F'}_{i})).
    \label{equ:sce}
\end{equation}
To ensure the preservation of global semantics, a regularization on $\Delta w$ is added, formulate as
\begin{equation}
    L_{\Delta} = \sum_i \|\Delta w_i\|_1 .
    \label{equ:norm_delta}
\end{equation}
Equipped with Style Code Enhancer, $E^3$-FaceNet can better embed fine-grained text information into the rendering process, as it refines and augments missing semantic information based on both the given text and the intermediate visual features while maintaining the image quality. 

\textbf{T3D face manipulation.} Notably, the proposed SCE also enables $E^3$-FaceNet to perform 3D manipulation. Specifically, we first obtain the predicted style code offsets, $w_{ori}$ and $w_{edit}$, corresponding to the original text prompt and manipulation instruction using Eq.\ref{equ:sce_ca1}-Eq.\ref{equ:sce}. Then a simple linear interpolation of two style code offsets is applied to refine the final style code offset $\Delta w$ by
\begin{equation}
    \Delta w = (1-\lambda) \Delta w_{ori} + \lambda \Delta w_{edit}
\end{equation}
where $\lambda \in [0,1]$ is a scale to control the editing degree. The updated style code is then used to influence the rendering process of the generated faces to edit the corresponding attribute while preserving the overall facial identity.

\subsection{Geometric Regularization in 3D Space}
\label{subsec:high_order_regularization}
Due to the lack of 3D supervision from the unposed 2D training dataset \cite{gu2021stylenerf}, a direct text-to-3D image mapping is prone to generating incorrect 3D geometries and shapes of human faces. To address this issue, we also equip $E^3$-FaceNet with a novel regularization on both basic geometric property and high-order geometric features.

Specifically, we focus on the points' \emph{3D location} and \emph{normal vector} corresponds to each pixel of the rendered feature map, as shown in Fig.\ref{fig:geometric}. These two geometric regularizations are designed to enforce smoothness and coherence among neighboring points, obtaining a more accurate representation of the 3D face's shape and facial attributes. 

 Formally, the estimated depth $\hat{d}$ of a ray $\mathbf{r}$ can be defined by
\begin{equation}
    \hat{d}(\mathbf{r})=\int_{t_n}^{t_f}T(t)\sigma_w(\mathbf{r}(t))tdt.
\end{equation}
Therefore, the 3D location $P_{(u,v)}(x,y,z)$ of pixel $p(u,v)$ in the world coordinate can be easily obtained via
\begin{equation}
   P_{(u,v)}(x,y,z) = P_c+\hat{d}(\mathbf{r_{(u,v)}})\cdot \mathbf{d},
   \label{equ:get_location}
\end{equation}
where $P_c$ is the camera coordinate and $\mathbf{d}$ is the normalized viewing direction. $\mathbf{d}$ can be easily obtained by the sampled camera pose $p$ and the pre-defined camera intrinsic matrix. Then the 3D location constraint is defined by
\begin{equation}
    L_{loc}=\sum_{(u,v)}\sum_{(i,j)\in\{-1,1\}^2}\|P_{(u+i,v+j)}-P_{(u,v)})\|_1.
    \label{equ:loc}
\end{equation}

In terms of the normal vector of $P$, previous works formulate it as $n=-\frac{\nabla_{\boldsymbol{x}}\sigma}{\|\nabla_{\boldsymbol{x}}\sigma\|}$ \cite{boss2021nerd,srinivasan2021nerv}. Since we can calculate the gradient $\nabla_{\boldsymbol{x}}\sigma$ directly, we turn to approximate $n$ by virtual normal \cite{yin2019enforcing} to reduce GPU memory and computation. 

Following \cite{klasing2009comparison} and \cite{yin2019enforcing}, we assume that local 3D points locate in the same plane of which the normal vector is the surface normal. Therefore, for each 3D point $P_{(u,v)}$, we select it two neighboring points,  \emph{e.g., $P_{(u+1,v)}$ and $P_{(u,v-1)}$}, to establish a plane and compute the normal vector by
\begin{equation}
    n_{(u,v)} = \frac{\overrightarrow{P_{(u,v)}P_{(u+1,v)}}\times\overrightarrow{P_{(u,v)}P_{(u,v-1)}}}{\left\|\overrightarrow{P_{(u,v)}P_{(u+1,v)}}\times\overrightarrow{P_{(u,v)}P_{(u,v-1)}}\right\|}.
\end{equation}

In this case, we perform regularization on the normal vectors, defined by
\begin{equation}
\resizebox{.86\hsize}{!}{$L_{normal}=\sum_{(u,v)}\sum_{(i,j)\in\{-1,1\}^2}\|n_{(u+i,v+j)}-n_{(u,v)}\|_1$}.
    \label{equ:normal}
\end{equation}

Conclusively, the final smooth regularization loss can be written by
\begin{equation}
    L_{reg} = L_{loc} + L_{normal}.
\label{equ:smooth}
\end{equation}

By integrating these regularizations, $E^3$-FaceNet significantly enhances face synthesis, achieving a more accurate 3D shape and improved fidelity.

\subsection{Objective Function}
 In terms of optimization, $E^3$-FaceNet adopts a non-saturating GAN objective \cite{goodfellow2014generative} with \emph{R1} regularization \cite{mescheder2018training} for stable training. The objective for the discriminator is
\begin{equation}
\begin{aligned}L_{D}=&-\frac{1}{2}[\mathbb{E}_{I\sim P_{dala}}logD(I)+\mathbb{E}_{\hat{I}\sim p_{G}}log(1-D(\hat{I}))\\&+\mathbb{E}_{{I\sim P_{dala}}}logD(I,T)+\mathbb{E}_{{\hat{I}\sim p_{G}}}log(1-D(\hat{I},T))]\\&+\gamma\mathbb{E}_{I\sim P_{dala}}[(\|\nabla D(I)\|^2+\|\nabla D(I,T)\|^2)].\\\end{aligned}    
\end{equation}

In terms of the generator, its objectives are 5-fold. First, $E^3$-FaceNet adopts the conditional loss \cite{mirza2014conditional} in adversarial training for text-guided generation, which is formulated as
\begin{equation}
    L_{adv}=-\frac{1}{2}[\mathbb{E}_{\hat{I}\sim p_{G}}logD(\hat{I})+\mathbb{E}_{\hat{I}\sim p_{G}}logD(\hat{I},T)].
\end{equation}
Meanwhile, NeRF-path regularization loss $L_{NeRF-path}$ \cite{gu2021stylenerf} is used to enforce 3D consistency. To achieve fast convergence, we also adopt a contrastive loss $L_{clip}$ that is defined by
\begin{equation}
    L_{clip}=-\sum_{i=1}^n\log\frac{\exp(E_T(T_i)\cdot E_I(\hat{I}_i))}{\sum_{j=1}^n\exp(E_T\cdot(T_i)\cdot E_I(\hat{I}_j))},
\end{equation}
where $T_i$ and $\hat{I}_i$ are the input text prompts and corresponding generated image. In this case, the overall objectives are:
\begin{equation}
    L_{_G}=L_{adv}+L_{reg}+L_{\Delta}+L_{clip}+\beta L_{NeRF-path},
\end{equation}
where $\beta$ is a hyper-parameter, and $L_{\Delta}$ is the style regularization term defined in Eq.\ref{equ:norm_delta}.

%% file: sec/4_exp.tex
\section{Experiment}
\label{sec:expe}
\subsection{Experimental Settings}
\textbf{Datasets} The benchmarks used in this paper include MMCelebA \cite{xia2021tedigan} , \emph{CelebAText-HQ} \cite{sun2021multi} and \emph{FFHQ-Text} \cite{zhou2021generative}. MMCelebA consists of 30,000 face images while CelebAText-HQ has 15,015 face images, each of which corresponds to 10 different text descriptions. FFHQ-Text is a small-scale face image dataset with diverse facial attributes. This dataset contains 760 female FFHQ faces with 9 descriptions for each image. We only use the training split of MMCelebA to train our $E^3$-FaceNet and evaluate it on the test split along with other 2D methods trained on the same dataset. In terms of 3D methods, as none of the available models can be trained on MMCelebA, we conducted zero-shot validation on the CelebAText-HQ dataset and FFHQ-Text for a fair comparison, which can also assess the generalization.

\noindent\textbf{Metrics} Following the settings of \cite{yu2023towards,wu2023high, aneja2023clipface}, we evaluate the generation quality of our method and the compared methods from three aspects: (1) \textit{Image-Quality}, \emph{i.e.}, the reality and fidelity of the rendered 2D face images, using \emph{Fréchet Inception Distance} (\textbf{FID}) \cite{heusel2017gans} and \emph{Kernel Inception Distance} (\textbf{KID}) \cite{binkowski2018demystifying} metrics, (2) \textit{Semantic-Alignment} (\textbf{SA}), \emph{i.e.}, using the text-image similarity of CLIP to measure whether the generated face align with the given description, and (3) \textit{Multi-View Identity Consistency} (\textbf{MVIC}), \emph{i.e.}, calculating the mean \emph{Arcface cosine similarity} \cite{schroff2015facenet} scores between the face images synthesized from the same 3D face but rendered from different camera poses. For better evaluations, we also conduct a user study from the perspectives of \emph{Identity Preservation} (\textbf{IP}), \emph{Semantic-Alignment} (\textbf{SA}) and \emph{Editing-Quality} (\textbf{EQ}). For more details refer to our appendix. 

\begin{table}[t]
\renewcommand{\arraystretch}{1.2}
\centering
\vspace{-4mm}
\caption{Comparison with existing T3D Face methods. Our $E^3$-FaceNet not only has better performance but also a faster speed. 
}
\footnotesize{
\setlength{\tabcolsep}{0.8mm}{
\begin{tabular}{c|cc|cc|c}
\toprule
\multirow{2}{*}{\textbf{Method}}& \multicolumn{2}{c|}{\textbf{CelebAText}} & \multicolumn{2}{c|}{\textbf{FFHQ-Text}} &   \textbf{Inference} \tablefootnote{We test the speed of five-view generation with one V100 GPU.}\\
\cline{2-5}
&\textbf{MVIC}($\uparrow$)&\textbf{SA} ($\uparrow$)& \textbf{MVIC}($\uparrow$)&\textbf{SA}($\uparrow$)&\textbf{Time}\\
\midrule
Latent3D &0.7835&\textbf{0.2688}&0.7845&0.2758&301.92s\\
Describe3D  &0.7973&0.2449&0.8027&0.2402&22.93s\\
\midrule
$E^3$-FaceNet&\textbf{0.8521}&0.2623&\textbf{0.8467}&\textbf{0.2778}&\textbf{0.64s}\\
 \bottomrule
\end{tabular}%
}}
\vspace{-3.8mm}
\label{tab:t2f3d}
\end{table}

\begin{figure*}[!h]
\centering
    \includegraphics[width=1.\textwidth]{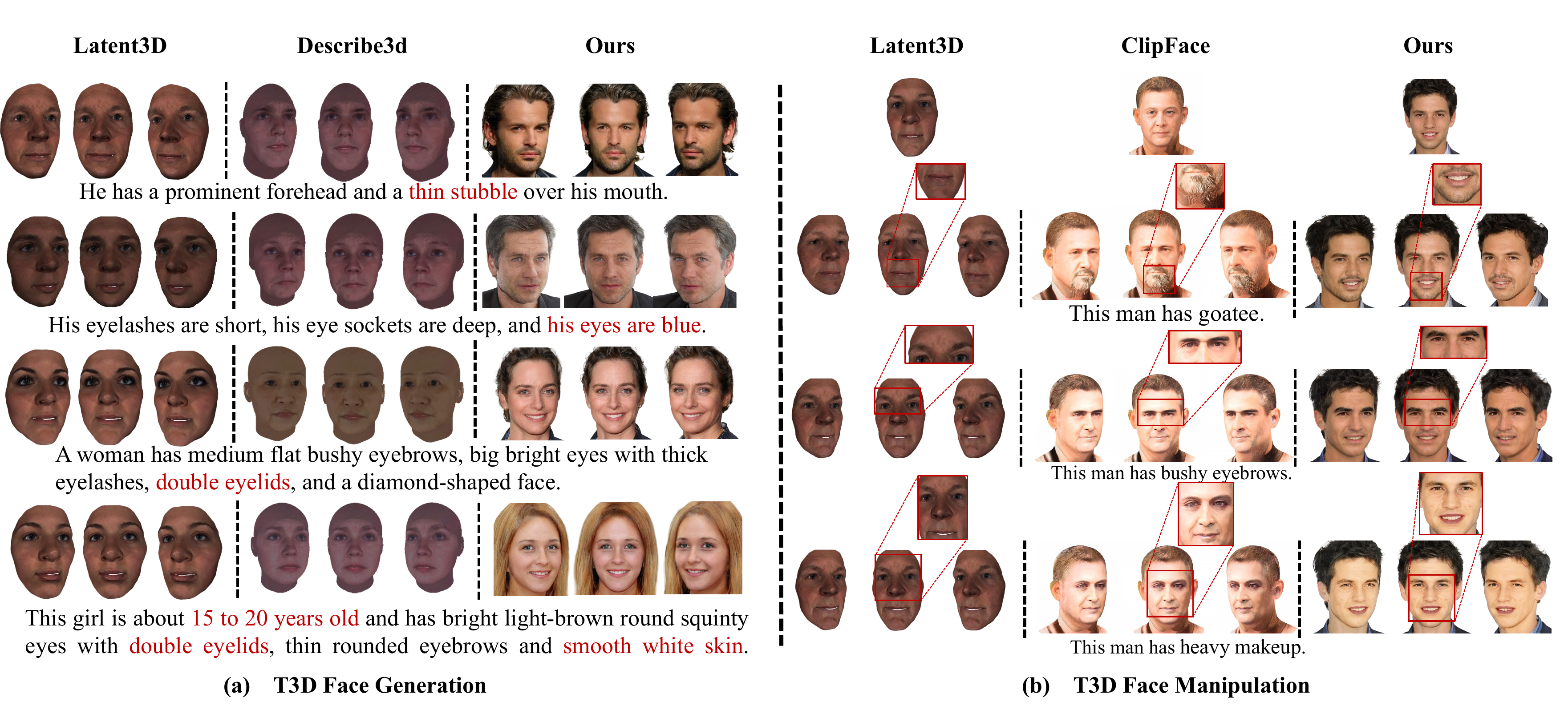}
    \vspace{-8.5mm}
\caption{Comparison with T3D Face generation and manipulation methods. $E^3$-FaceNet can generate face images of better quality than the compared methods, while well aligning the generation and editing prompts\protect\footnotemark[2]. Its inference speed is also much faster.}
\vspace{-0.5em}
\label{fig:3D}
\end{figure*}

\noindent\textbf{Implementation Details} In this paper, we use StyleNeRF \cite{gu2021stylenerf} as our base model. Similarly, we adopt NeRF++ \cite{zhang2020nerf++}, a variant of NeRF, as the backbone. We follow the settings of \cite{gu2021stylenerf} and use NeRF++ to produce a feature map at 32x32 resolution, which is then progressively upsampled to 512x512 resolution. We trained our $E^3$-FaceNet with a learning rate of 2.5e-4 for both the generator and discriminator. To expedite convergence, we initialized the model with pre-trained weights of StyleNeRF, which were trained on the FFHQ dataset at a resolution of 512. We set $\gamma=0.5$ and $\beta=0.2$. The entire process for our model takes approximately about 5 days on 4 V100 GPUs.

\subsection{Experimental Results}

\noindent\textbf{Comparison with T3D Face methods.} We first quantitatively compare our $E^3$-FaceNet with two strong T3D Face methods, namely Latent3D \cite{canfes2023text} and Describe3D \cite{wu2023high} on CelebAText \cite{sun2021multi} and FFHQ-Text \cite{zhou2021generative}, of which results are given in Tab.\ref{tab:t2f3d}. From this table, we can first find that $E^3$-FaceNet greatly outperforms the other two methods under the metric of MVIC, \emph{e.g.}, +6.87\% on CelebAText, showing great advantages in terms of multi-view identity preservation. Meanwhile, the performance of SA is also outstanding, which is slightly worse than Latent3D on CelebAText but obviously better on FFHQ. These results suggest that $E^3$-FaceNet can achieve excellent multi-view semantic alignment. Moreover, retaining high performance, our $E^3$-FaceNet also speeds up the five-view generations by orders of magnitudes. For example, compared with Latent3D and Describe3D, the inference speed of $E^3$-FaceNet is 471.75 $\times$ and 35.83 $\times$ faster. Notably, $E^3$-FaceNet achieves such performance under the setting of zero-shot generation. 
\begin{table}[!t]
\centering
\vspace{-3mm}
\caption{Comparison with existing T3D face manipulation. The inference time is for five-view generation.}
\footnotesize{
\setlength{\tabcolsep}{2.3mm}{
\begin{tabular}{c|ccc|c}
\toprule
\textbf{Method}& \textbf{IP} ($\uparrow$) & \textbf{SA}($\uparrow$) & \textbf{EQ}($\uparrow$) & \textbf{Inference Time}($\downarrow$) \\
\midrule
Latent3D  & 21.92&2.8 &  11.60  & 304.44s\\ 
ClipFace  & 8.16& 44.56&  12.48 & 17min + 1.83s\\ 
\midrule
Ours    &\textbf{69.92}  &\textbf{52.64} & \textbf{75.92} & \textbf{0.89s}\\ 
 \bottomrule
\end{tabular}%
}}
\vspace{-5mm}
\label{tab:editing}
\end{table}

\noindent\textbf{Text-Driven 3D Face Manipulation.}
To examine the editing ability of $E^3$-FaceNet, we also compare it with Latent3D and ClipFace \cite{aneja2023clipface} in Tab.\ref{tab:editing}. Without available evaluations, we conduct a user study for this comparison. We can first observe that the identity preservation and the editing quality of $E^3$-FaceNet are much superior to the compared methods, \emph{e.g.}, 69.92 \emph{v.s.} 21.92 of Latent3D on IP and 75.92 \emph{v.s.} 12.48 of ClipFace on EQ. Besides, $E^3$-FaceNet also exhibits strong cross-modal semantic consistency, \emph{e.g.}, 52.64 \emph{v.s.} 44.56 of ClipFace on SA.  In addition, the efficiency of $E^3$-FaceNet is still much better than the others. For instance, ClipFace requires about 17 \emph{mins} to train the texture mappers with another 1.8 seconds for rendering. In contrast, $E^3$-FaceNet only takes 0.89 seconds for the five-view editing. 

\begin{table}[!t]
\vspace{-2.8mm}
\centering
\caption{Comparison with T2D face methods on MM-CelebA.
}
\footnotesize{
\setlength{\tabcolsep}{2.8mm}{
\begin{tabular}{c|cc|c}
\toprule
\textbf{Method}            & \textbf{FID} ($\downarrow$) & \textbf{KID} ($\downarrow$) &   \textbf{CLIP-Score} ($\uparrow$) \\
\midrule
StyleCLIP  & 53.38& 49.89 &  0.2452    \\ 
TediGAN       & 47.61& 45.63&  \textbf{0.2865}  \\ 
\midrule
OpenFaceGAN        & 20.72 & 12.58&   0.2590  \\ 
PixelFace & 16.63 & 8.65&   0.2583    \\ 
\midrule
Ours              &\textbf{12.46}&  \textbf{5.84} &  0.2770   \\
 \bottomrule
\end{tabular}%
}}
\vspace{-6mm}
\label{tab:t2f2d}
\end{table}
\footnotetext[2]{The backgrounds are removed for better comparison.}

\begin{table*}[!t]
\renewcommand{\arraystretch}{1.2}
\centering
\vspace{-2.5mm}
\caption{Ablation study of $E^3$-FaceNet. $L_{reg}$ is the proposed \emph{geometric regularization} and SCE is Style Code Enhancer.  
}
\footnotesize{
\setlength{\tabcolsep}{3.5mm}{
\begin{tabular}{c|ccc|cc|cc}
\toprule
 \multirow{2}{*}{\textbf{Setting}}& \multicolumn{3}{c|}{\textbf{MMCelebA}}& \multicolumn{2}{c|}{\textbf{CelebAText}} & \multicolumn{2}{c}{\textbf{FFHQ-Text}} \\
\cline{2-8}
&\textbf{FID}($\downarrow$)&\textbf{KID}($\downarrow$)&\textbf{CLIP-Score}($\uparrow$)&\textbf{MVIC}($\uparrow$)&\textbf{SA}($\uparrow$)&\textbf{MVIC}($\uparrow$)&\textbf{SA}($\uparrow$)\\
\hline
w/o $L_{reg}$ \& SCE &13.44 &	6.11 &	0.2622& 	0.7960& 	0.2499 &	0.7873& 	0.2681    \\
\midrule
Only $L_{loc}$ in Eq.\ref{equ:loc} & 12.87& 	6.16& 	0.2657 &	0.8431 &	0.2547 &	0.8313 &	0.2711 \\
Only $L_{normal}$ in Eq.\ref{equ:normal} & 12.78 &	5.94 &	0.2703 &	0.8307& 	0.2570 &	0.8278 &	0.2730  \\
$L_{reg}$ in Eq.\ref{equ:smooth} & 12.72 &	5.92& 	0.2652 &	\textbf{0.8560} &	0.2540& 	\textbf{0.8511}& 	0.2704  \\ 
\midrule
$L_{reg}$+SCE& \textbf{12.46} &	\textbf{5.83}& 	\textbf{0.2770}& 	0.8521& 	\textbf{0.2623} &	0.8467 &	\textbf{0.2778}  \\
 \bottomrule
\end{tabular}%
}
}
\label{tab:ablation}
\end{table*}

\noindent\textbf{Comparison with T2D Face methods.} To further examine the generation ability of $E^3$-FaceNet, we also compare it with a set of strong T2D Face methods on MMCelebA \cite{xia2021tedigan,patashnik2021styleclip,peng2022learning,peng2022towards}, and the result is shown in Tab.\ref{tab:t2f2d}. Theoretically, 2D face generation is relatively easier than 3D one \cite{toshpulatov2021generative}. However, $E^3$-FaceNet also achieves outstanding performance on FID and KID for 2D image quality, indicating that $E^3$-FaceNet has better perceptual quality. In terms of CLIP-Score for semantic alignment, $E^3$-FaceNet is slightly worse than TediGAN \cite{xia2021tedigan}, which requires an example-dependent optimization, \emph{i.e.}, 200 steps for each text. Compared with the end-to-end 2D methods, \emph{i.e.}, OpenFace \cite{peng2022towards} and PixelFace \cite{peng2022learning}, $E^3$-FaceNet is still better in semantic consistency, well confirming the merits of our SCE. Overall, $E^3$-FaceNet also exhibits a strong single-view generation ability, even better than existing SOTA T2D Face methods.

\noindent\textbf{Ablation study.} We ablate different designs of $E^3$-FaceNet on four benchmarks in Tab.\ref{tab:ablation}. Compared with the plain baseline,  \emph{i.e.}, without $L_{reg}$ and SCE, the location loss in Eq.\ref{equ:loc} or the normal smoothness in Eq.\ref{equ:normal} can help $E^3$-FaceNet to generate smooth surfaces and greatly improve  MVIC score. Meanwhile, their combination, \emph{i.e.}, \emph{$L_{reg}$}, achieves better performance. Besides, the introduction of SCE can inject finer-grained text information into the generation process, as evidenced by achieving the highest SA. 
\begin{figure}[t]
    \vspace{-2em}
\includegraphics[width=.49\textwidth]{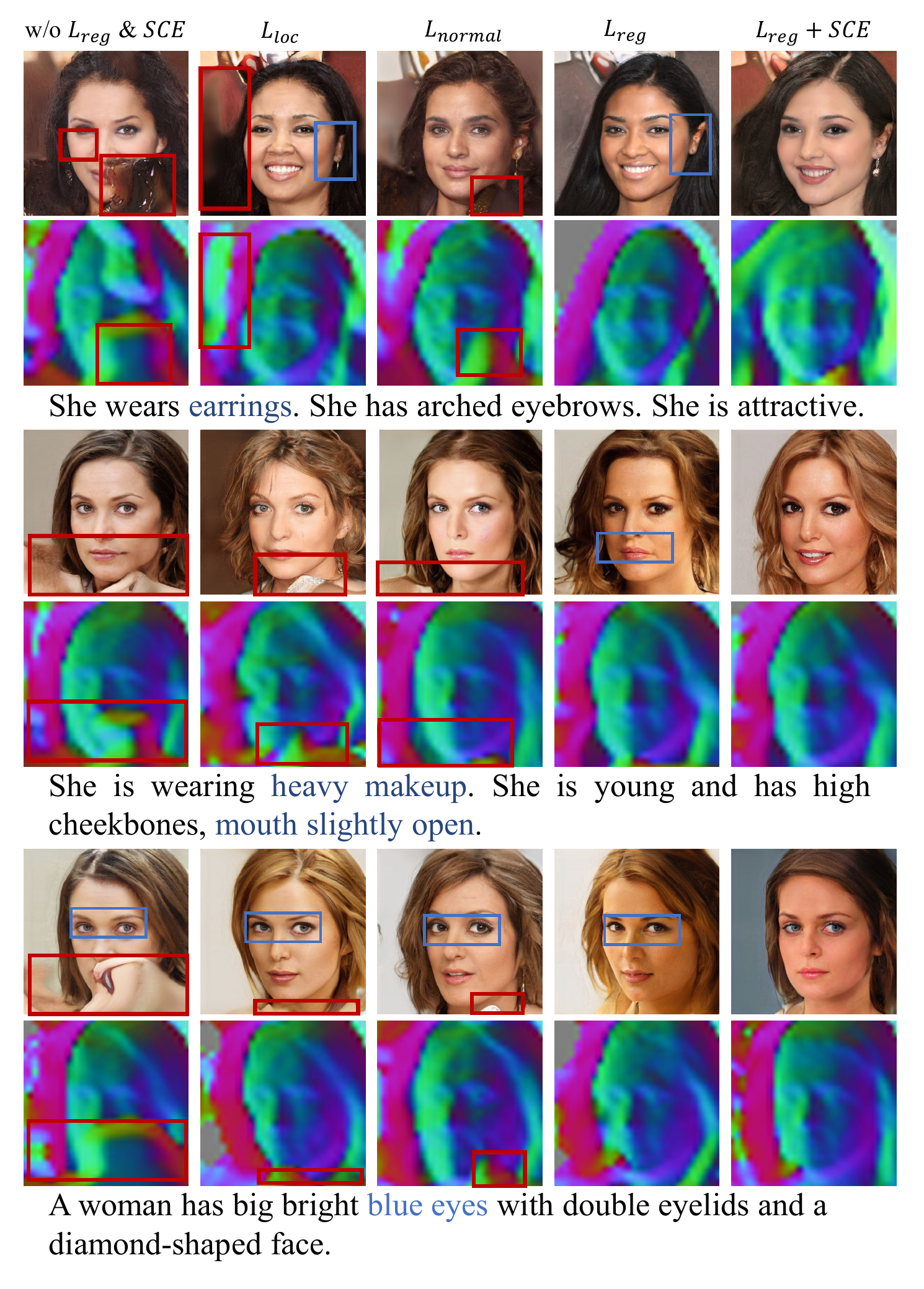}
    \vspace{-2em}
    \caption{The visualizations of different settings of $E^3$-FaceNet. The bottom images are the correspondent feature map}
    \label{fig:ablation_div}
\vspace{-1.2em}
\end{figure}

\noindent\textbf{Visualizations of Text-to-3D Face Generation.} We visualize the T3D face generations of $E^3$-FaceNet, Latent3D and Describe3D in Fig.\ref{fig:3D} (a). As can be seen, $E^3$-FaceNet excels in synthesizing high-quality images with better multi-view consistency, which also well aligns with the given prompts. Meanwhile, Latent3D and Describe3D are hard to reconstruct faces with finer-grained details, as exemplified by the prompt ``\emph{thin stubble}'' in Fig.\ref{fig:3D} (a). Also, these two methods fall short of generating 3D faces with diverse facial attributes, such as hair or ears, lacking the ability to incorporate such elements into the synthesized faces.

\noindent\textbf{Visualizations of Text-Driven 3D Face Manipulation.} We also present the editing results of $E^3$-FaceNet, Latent3D and ClipFace in Fig.\ref{fig:3D} (b). It can be observed that Latent3D struggles to respond to editing instructions, likely due to its heavy reliance on the initially generated face \cite{canfes2023text}. For instance, neither the ``\emph{goatee}'' nor the ``\emph{heavy makeup}'' are generated in its editions. ClipFace is slightly better and aware of the text semantics. However, without geometry optimization, ClipFace synthesizes unnatural texture maps and barely achieves multi-view consistency, \emph{e.g.,} ``\emph{goatee} ''. The identity also changes significantly. In contrast, $E^3$-FaceNet can well respond to the text semantics while retaining high quality and multi-view consistency.

\noindent\textbf{The impacts of different designs in $E^3$-FaceNet.} We also visualize the impacts of each design in Fig.\ref{fig:ablation_div}. Without the geometric regularization, the model tends to generate faces of artifacts, which is most severe under the base settings. In contrast, this loss term can help $E^3$-FaceNet synthesize 3D faces with clear contour. Additionally, with SCE, finer-grained semantics can be seamlessly injected. For example, the model successfully generates earrings, while other settings fail to capture this attribute. 

\begin{figure}[t]
    \vspace{-4mm}
    \includegraphics[width=0.48\textwidth]{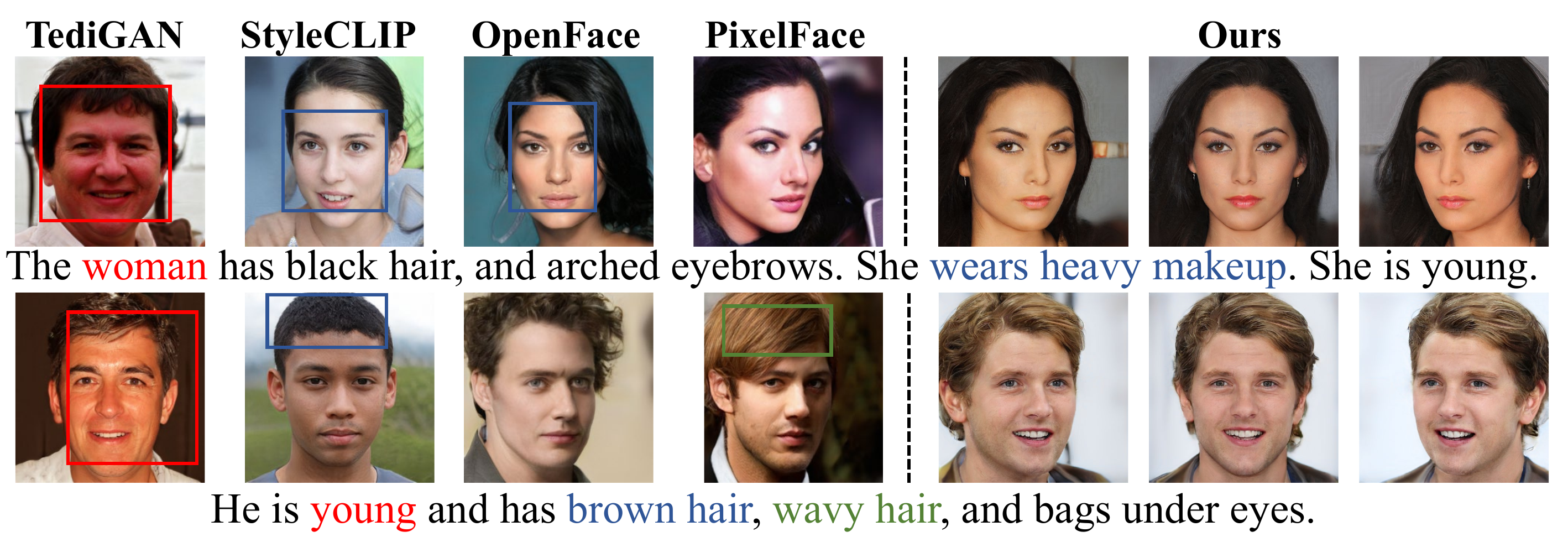}
    \vspace{-2.5em}
    \caption{The comparison with text-to-2D face methods. $E^3$-FaceNet also excels in image quality and semantic consistency.}
    \label{fig:compare_2d}
    \vspace{-1.em}
\end{figure}
\noindent\textbf{Visualizations of Text-to-2D Face Generation.} In Fig.\ref{fig:compare_2d}, we also visually compare the single-view generation of $E^3$-FaceNet with strong T2D Face methods \cite{xia2021tedigan,patashnik2021styleclip,peng2022learning,peng2022towards}. As can be seen, $E^3$-FaceNet can achieve better semantic alignment than these 2D methods, \emph{e.g.}, gender, age, or hairstyles. Compared with the end-to-end 2D methods, \emph{i.e.}, OpenFace and PixelFace, $E^3$-FaceNet also presents better generation qualities of different views. These results well confirm the excellent generation ability of $E^3$-FaceNet again. 

%% file: sec/5_conclusion.tex
\section{Conclusion}
\label{sec:conclusion}
In this paper, we propose a novel and efficient $E^3$-FaceNet for fast and accurate T3D face generation and manipulation. $E^3$-FaceNet follows the principle of direct cross-modal mapping, and introduces a novel Style Code Enhancer for semantic-aligned rendering and an innovative Geometric Regularization to ensure multi-view consistency. The experiment results demonstrate that $E^3$-FaceNet achieves higher-fidelity generation and better semantic consistency than a set of compared T2D and T3D Face methods, and also speed up T3D face generation by orders of magnitudes.

%% file: sec/x_appendix.tex
\setcounter{footnote}{0}
In this supplement, we first introduce the comparison methods and their experimental setup in \cref{sec:comparison}. Subsequently, we provide a comprehensive description of the evaluation metrics in \cref{sec:evaluation}. We also present additional visualization results in \cref{sec:additional_comparison} and \cref{sec:additional_results}.

\section{Introduction of Comparison Methods}
\label{sec:comparison}
In this section, we introduce the comparison methods used in the experiment and their implementation settings.

\noindent\textbf{Latent3D} \cite{canfes2023text} builds upon the TB-GAN \cite{gecer2020synthesizing}, a generative model that inputs one-hot encoded facial expressions along with a random noise vector, and then outputs shape, shape-normal, and texture images. Typically, Latent3D optimizes the offset $\Delta c$ for the intermediate layer $\mathbf{c}$ of TB-GAN, directing how the target attributes, as specified by the text prompt, are enhanced. It employs a combination of a CLIP-based loss, an identity loss and an $L2$ loss as
\begin{equation}
\arg\min_{\Delta\mathbf{c}\in\mathcal{C}}\mathcal{L}_{\mathrm{CLIP}}+\lambda_{\mathrm{ID}}\mathcal{L}_{\mathrm{ID}}+\lambda_{\mathrm{L2}}\mathcal{L}_{\mathrm{L2}},
\end{equation}
where $\lambda_{ID}$ and $\lambda_{L2}$ are the hyper-parameters of $\mathcal{L}_{ID}$ and $\mathcal{L}_{L2}$, respectively. We use the official implement code \footnote{\url{https://github.com/catlab-team/latent3D_code}} for 3D face generation and manipulation. Following their suggestions, we set $\lambda_{ID}=0.01$, $\lambda_{L2}=0.0$ with a learning rate of 0.01 for 100 epochs per text.

\noindent\textbf{Describe3D} \cite{wu2023high} builds a dataset comprising 1,627 3D face models, each annotated with 25 labels representing facial attributes. This dataset is then used for training the entire generation pipeline, which consists of three stages: text parsing, concrete synthesis and abstract synthesis. The text parser is used to encode the input natural language into a descriptive code $d$, which is then divided into shape-related code $d_S$ and texture-related code $d_T$. The shape-related code $d_S$ is employed to predict 3DMM \cite{blanz2023morphable} parameters, while the texture-related code $d_T$ serves as a conditional input for StyleGAN2 \cite{karras2020analyzing} to generate texture maps. This process is known as concrete synthesis. To boost performance, Describe3D incorporates additional optimization to finetune both the 3DMM parameters and texture parameters during the abstract synthesis stage. We use the official code \footnote{\url{https://github.com/zhuhao-nju/describe3d}} and the released model to generate 3D faces.

\noindent\textbf{ClipFace} \cite{aneja2023clipface} also leverages the geometric expressiveness of 3D morphable models but introduces a self-supervised generative model to synthesize textures and expression parameters for the morphable model. Given a textured mesh with texture code $\mathbf{w_{init}}=\{\mathbf{w_{init}^{1}},...,\mathbf{w_{init}^{18}}\}\in \mathbf{R}^{512 \times 18}$, Clipface predict the offsets formulated as:
\begin{equation}
    \mathbf{w}_{\mathrm{delta}}^*,\boldsymbol{\psi}_{\mathrm{delta}}^*=\underset{\mathrm{Wdelta}}{\operatorname*{\arg\min}}\mathcal{L}_{\mathrm{delta}},
\end{equation}
where $\mathbf{w}_{\mathrm{delta}}$ and $\boldsymbol{\psi}_{\mathrm{delta}}$ are the optimized offsets for texture and expression, respectively.  $\mathcal{L}_{\mathrm{delta}}$ is the full training loss containing a CLIP-based loss for enhancing attributes depicted in the prompt and a regularization term for the facial expression. In practice, they employ a 4-layer MLP architecture for the mappers and we use the code from their official implement \footnote{\url{https://github.com/shivangi-aneja/ClipFace}} in the experiments.

\noindent\textbf{TediGAN} \cite{xia2021tedigan} can perform text-guided 2D image generation by optimizing a random sampled noise $z*$ via
\begin{gather}
    \begin{aligned}\mathbf{z}^*=\arg\min_\mathbf{z}||\mathbf{x}-G(\mathbf{z})||_2^2+\lambda_1^{\prime}||F(\mathbf{x})-F(G(\mathbf{z}))||_2^2\\+\lambda_2^{\prime}||\mathbf{z}-E_v(G(\mathbf{z}))||_2^2+\lambda_3'\mathrm{L}_{CLIP},\end{aligned}\\
    \mathrm{L}_{CLIP}=1-\langle E_T(T),E_I(G(z))\rangle,
\end{gather}
where $x$ is the original image of $z$, $G$ is the generator, $F$ is VGG network \cite{simonyan2014very}. $\lambda_1^{\prime}$, $\lambda_2^{\prime}$ and $\lambda_3^{\prime}$ are the loss weights corresponding to the perceptual loss, regularization term and CLIP loss, respectively. $E_v$ is the introduced inversion encoder, $E_T$ and $E_I$ are the CLIP text encoder and image encoder, respectively, $\langle \cdot, \cdot \rangle$ is the cosine similarity. Following the instructions of the official implement \footnote{\url{https://github.com/IIGROUP/TediGAN}}, we set $\lambda_1=5e-5$, $\lambda_2=2.0$ and $\lambda_3=1.0$. The learning rate is set to 1e-2 and the number of optimization steps for each instruction is 200.

\noindent\textbf{StyleCLIP} \cite{patashnik2021styleclip} is another text-guided image generation and manipulation method concurrent to TediGAN, which turns to optimize the latent code $w_s$ via
\begin{gather}
    \arg\min_{w\in\mathcal{W}+}1-\langle E_T(T),E_I(G(z))\rangle.
\end{gather}
 We use the official code \footnote{\url{https://github.com/orpatashnik/StyleCLIP}} and the learning rate is set to 0.1 with 200 optimization steps.
 
\noindent\textbf{OpenFace} \cite{peng2022towards} is a learning-based method designed for 2D text-to-face generation task. It first constructs an effective multi-modal latent space and then directly maps a text description to a latent code. This code is then fed to a StyleGAN architecture and can perform text-guided face generation, combination and manipulation. We follow the official code \footnote{\url{https://github.com/pengjunn/OpenFace}} and use the released model weights to generate single-view text-to-2D face.

\noindent\textbf{PixelFace} \cite{peng2022learning} employs pixel regression for face generation, wherein each pixel value is predicted based on the latent code and initialized pixel embeddings. The pixel embeddings are constructed using Fourier features and coordinate embeddings. To incorporate text features, PixelFace proposes a cross-modal dependency based dynamic parameter generation module to generate dynamic knowledge for pixel synthesis. We use the official code \footnote{\url{https://github.com/pengjunn/PixelFace}} and the released model for single-view generation.

\section{Evaluation Metric}
We utilize both 2D and 3D metrics to provide a comprehensive validation of our $E^3$-FaceNet. When comparing with 3D methods, we randomly sample 200 text descriptions for each dataset, \emph{i.e.,} \emph{CelebAText-HQ} \cite{sun2021multi} and \emph{FFHQ-Text} \cite{zhou2021generative}. For each description, we generate a corresponding 3D face, then render it from five different poses. The evaluation scores for the 3D metrics are calculated inside each group, then the average score across all 200 groups is obtained. As the texts in CelebAText and FFHQ-Text are quite different from those in the Multi-Modal CelebA-HQ dataset, they can be used for cross-dataset experiments. Note that in this comparison, all the models perform zero-shot generation, thus the generalization and robustness can be effectively assessed.  
When comparing with 2D methods, we randomly select 30,000 descriptions from \emph{MMCelebA} \cite{xia2021tedigan} and perform single-view generation for each text prompt. In this part, we compare $E^3$-FaceNet with a bunch of strong T2D Face generation methods trained on this dataset.
\label{sec:evaluation}
\subsection{FID \& KID}
\emph{Fréchet Inception Distance} (\textbf{FID}) \cite{heusel2017gans} and \emph{Kernel Inception Distance} (\textbf{KID}) \cite{binkowski2018demystifying} are two predominant metrics for assessing the quality of generated images. FID is defined by the Fréchet distance between the feature from the real and generated images, which can be formulated as
\begin{equation}
    FID=||\mu_r-\mu_g||^2+Tr(\Sigma_r+\Sigma_g-2(\Sigma_r\Sigma_g)^{1/2})
\end{equation}
where $Tr$ is the trace and $(\mu_r,\Sigma_r)$ and $(\mu_g,\Sigma_g)$ are the mean and covariance obtained from the real images and generated images feature, respectively. We use Inception-v3 \cite{szegedy2016rethinking} to extract the image feature and lower FID indicates better perceptual quality.

KID can be viewed as a Maximum Mean Discrepancy (MMD) \cite{binkowski2018demystifying} directly on input images with the kernel
\begin{gather}
    K(\mathcal{I}_r,\mathcal{I}_g) = k(\theta(\mathcal{I}_r),\theta(\mathcal{I}_g))\\
    k(x,y) = (\frac{1}{d}x^Ty+1)^3
\end{gather}
where $\mathcal{I}_r$ and $\mathcal{I}_g$ are the real and generated images, respectively, and $\theta$ is the function mapping images to Inception representations. In the main paper, we report KID$\times 1000$ and a lower score indicates better generation quality.

\subsection{Multi-View Identity Consistency}
 In line with previous studies \cite{wu2023high,yu2023towards}, we employ ArcFace \cite{schroff2015facenet} to extract face features from the rendered images. We calculate the average cosine similarities between any two out of five images inside each group, and the multi-view identity consistency (MVIC) is obtained by computing the mean similarity score across 200 groups for each method on each dataset. A higher MVIC score indicates a greater level of multi-view consistency.
\subsection{Semantic-Alignment}
To evaluate the semantic alignment between text descriptions and the generated faces, we employ CLIP \cite{radford2021learning} to extract features from both the rendered images and text descriptions. We then measure their similarity in the CLIP space. For 3D methods, we calculate the average CLIP similarity for each group and compute the overall average CLIP similarity across all generated results. In the case of 2D methods focused on single-view generation, we calculate the CLIP score between each text description and its corresponding generated image, and then derive the mean score for this metric. A higher score indicates better semantic alignment.
\begin{figure*}[!t]
    \centering
    \includegraphics[width=0.9\linewidth]{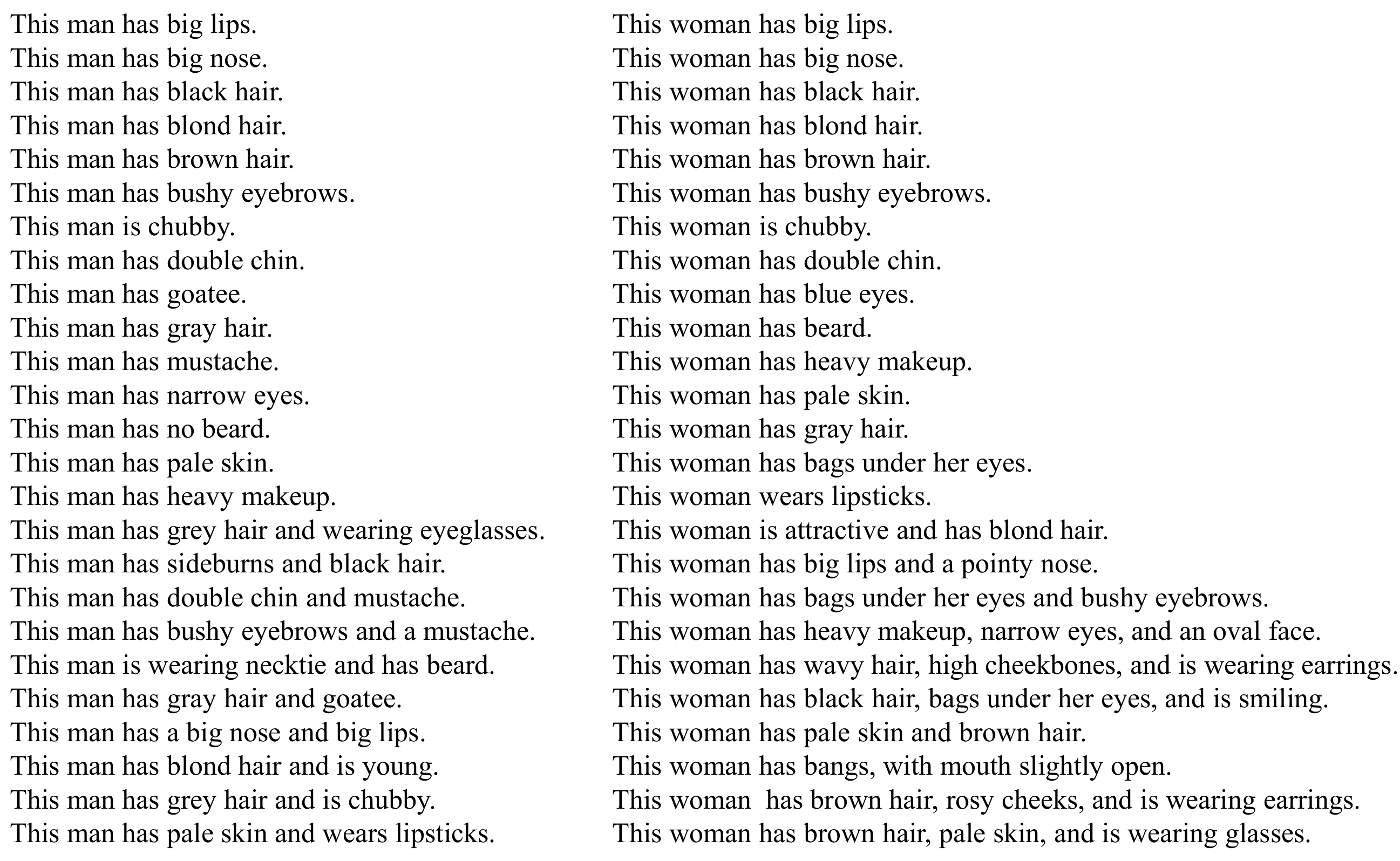}
    \caption{Editing instructions for text-driven 3D face manipulation comparison. We collect a total of 25 prompts for both male (left) and female (right) faces, including instructions for single-attribute editing and multi-attribute editing. }
    \label{fig:edit_prompt}
    \vspace{-5mm}
\end{figure*}

\subsection{User Study}
There is currently no standardized metric specifically designed for evaluating face manipulation. Previous studies, such as ClipFace \cite{aneja2023clipface}, rely on FID which is widely used for quantitative comparison in image generation tasks. However, comparing the two methods we are evaluating, namely Latent3D \cite{canfes2023text} and ClipFace \cite{aneja2023clipface}, is problematic because they are trained on different datasets. Using the aforementioned metrics under these circumstances could lead to unfair comparisons. Additionally, obtaining a reliable FID score requires a large collection of generated images. Given that both methods under evaluation rely on inference-time or example-dependent optimization, producing a sufficient quantity of edited faces is a time-intensive task. Therefore, to ensure a more accurate and effective quantitative comparison, we also carry out an additional user study, following \cite{xia2021tedigan,yu2023towards}.

\subsubsection{Editing Instruction}
We manually collected a set of 50 editing instructions to perform face editing across various attributes, consisting of 25 instructions for male faces and 25 for female faces.  In each group, we create 15 prompts for single-attribute manipulation and 10 prompts for multi-attribute manipulation. Additionally, we include some unconventional edits, such as applying ``\emph{lipstick}'' to a man or adding ``\emph{beard}'' to a woman, to further assess the editing capabilities of each method. The editing prompts are presented in \cref{fig:edit_prompt}.

\begin{figure*}[!t]
\centering
\subfloat[Examples for Male Face Manipulation]{
\begin{minipage}[t]{0.5\textwidth}
  \centering
  \includegraphics[width=0.95\textwidth]{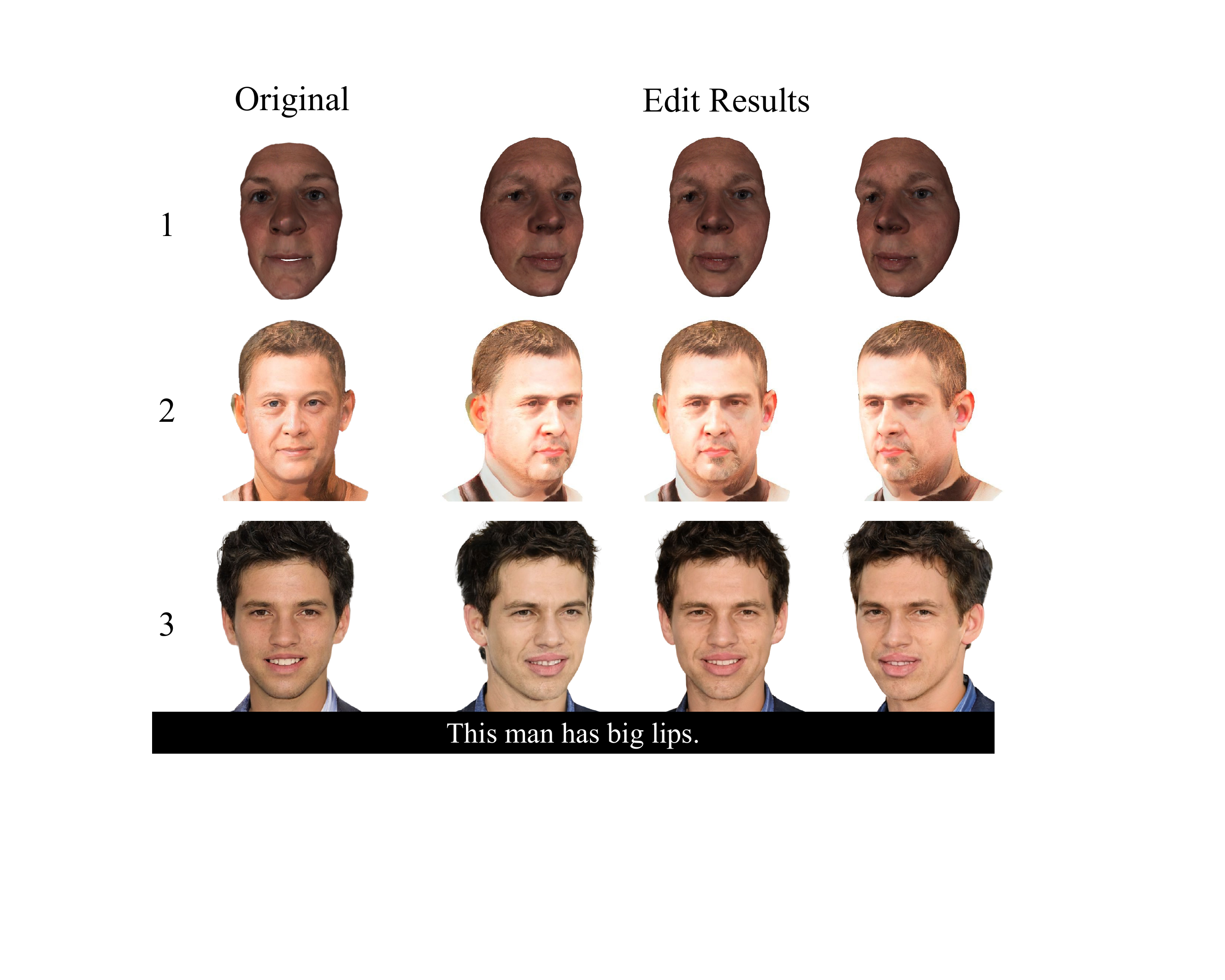}
  \end{minipage}
}
\subfloat[Examples for Female Face Manipulation]{
\begin{minipage}[t]{0.5\textwidth}
  \centering
  \includegraphics[width=0.95\textwidth]{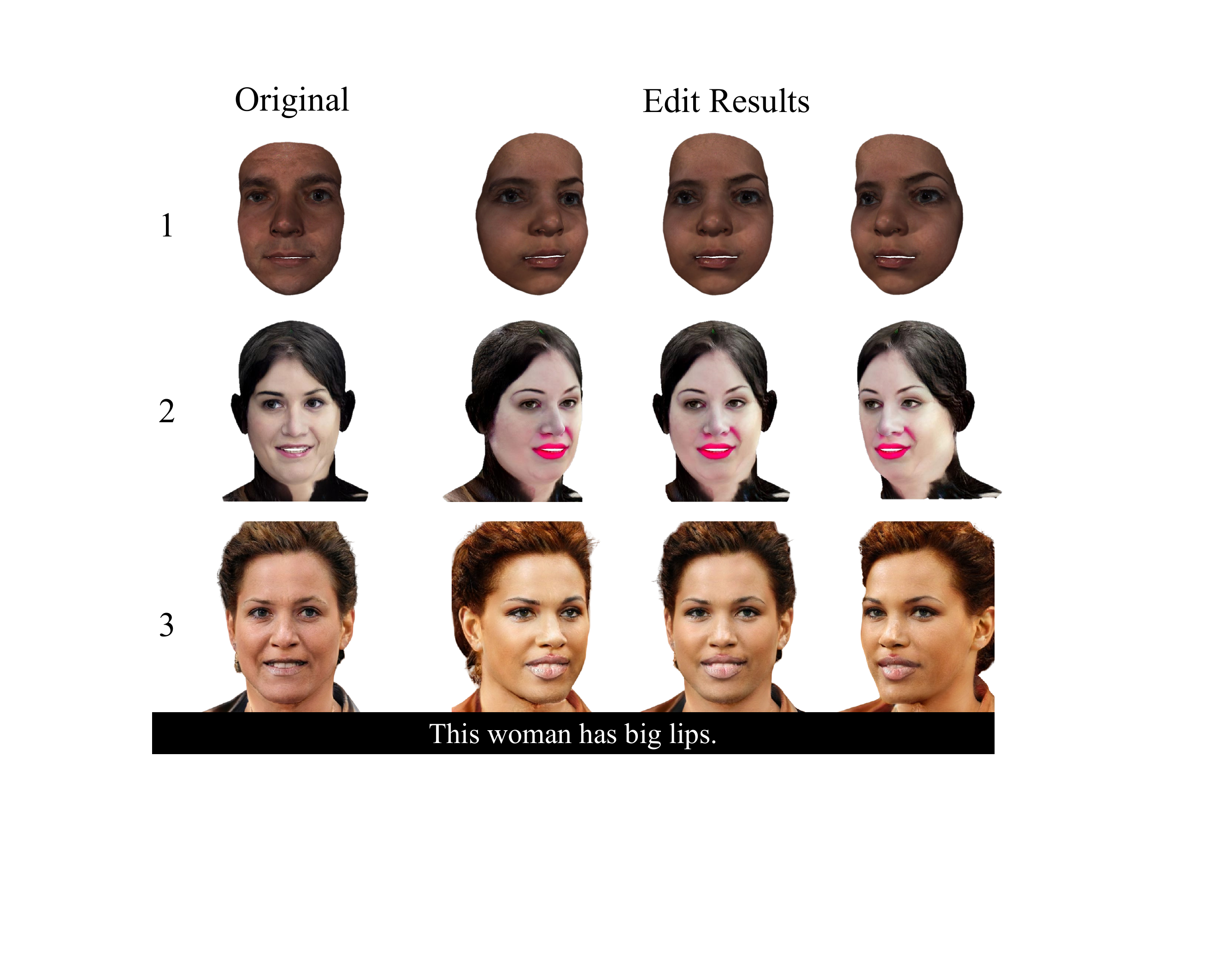}
  \end{minipage}
}
\caption{Examples of user study. On the left is the original face, followed by the edited results of the given prompt under each group. To ensure anonymity, we only present these generation results with indexes for recording purposes.}
\label{fig:edit_userstudy}  
\end{figure*}

\subsubsection{Detailed Experiment Settings}
For Latent3D, we initially generate a 3D face and then perform optimization based on the provided prompt. For ClipFace, we follow their templates and construct the editing instruction as \emph{``A photo of a male face with thick lips''} or \emph{``A photo of a male face with narrow eyes''}. We train specific mappers for each instruction and randomly select a textured female face and a male face for manipulation. Regarding $E^3$-FaceNet, we generate the initial 3D face using two prompts, \emph{i.e.,} \emph{``This is a man''} and \emph{``This is a woman''}. Then, we edit the generated face using the proposed Style Code Enhancer, with the default edit weight set to 0.5.
\begin{figure*}[!t]
    \centering
    \includegraphics[width=0.9\linewidth]{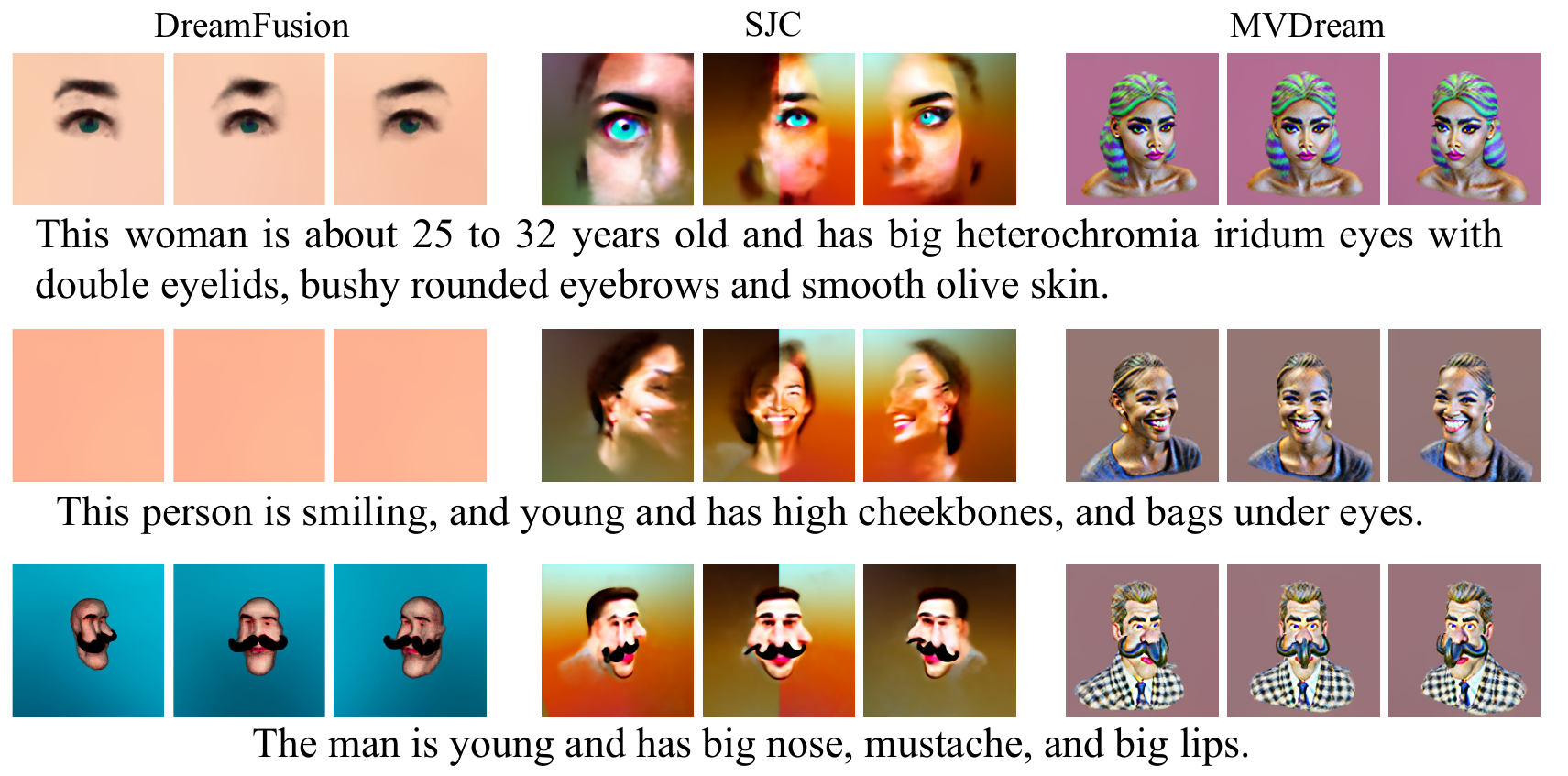}
    \caption{Results of generating 3D face of the current diffusion-distilled-based methods. Although these methods succeed in generating 3D objects, they cannot synthesize 3D human faces with high fidelity.}
    \label{fig:diff_nerf}
\end{figure*}
\subsubsection{Evaluation Criteria}
For each editing instruction, we render manipulated 3D faces from three different views: left, front, and right. Participants are presented with the original face and all editing results. We invite 25 individuals to assess the edited faces based on three criteria: (1E) \emph{Identity Preservation} (\textbf{IP}), selecting the edit result that best preserves the facial identity of the original face; (2) \emph{Semantic Alignment} (\textbf{SA}), selecting the edit result that most closely aligns with the given instructions; and (3) \emph{Editing Quality} (\textbf{EQ}), selecting the edit result with the highest perceptual quality. We calculated the percentage of times each method was chosen as the best performer by users, based on the previous mentioned criteria. As shown in \cref{fig:edit_userstudy}, the edit results are presented anonymously to the users. 

\section{Additional Comparisons}
\label{sec:additional_comparison}
Recently, researchers have been focused on transferring pre-trained 2D image-text diffusion models to synthesize 3D objects without using any 3D data \cite{poole2022dreamfusion,wang2023score,shi2023mvdream,wang2023prolificdreamer,lin2023magic3d}. While these methods have achieved competitive quality in 3D content creation, they fall short in generating photo-realistic human faces. Here we select three representative methods, \emph{i.e.,} Dreamfusion \cite{poole2022dreamfusion}, SJC \cite{wang2023score} and MVDream \cite{shi2023mvdream}, for 3D face generation.  However, as depicted in \Cref{fig:diff_nerf}, these methods struggle to generate high-quality 3D faces conditioned on the given input. Additionally, they all rely on test-time tuning, which significantly increases the inference time. For example, Dreamfusion takes approximately 40 minutes, and SJC takes around 20 minutes for inference on an RTX 3090 GPU. Therefore, these methods are not included in the comparison for T3D.

\begin{figure*}[!t]
    \centering
    \includegraphics[width=1.0\linewidth]{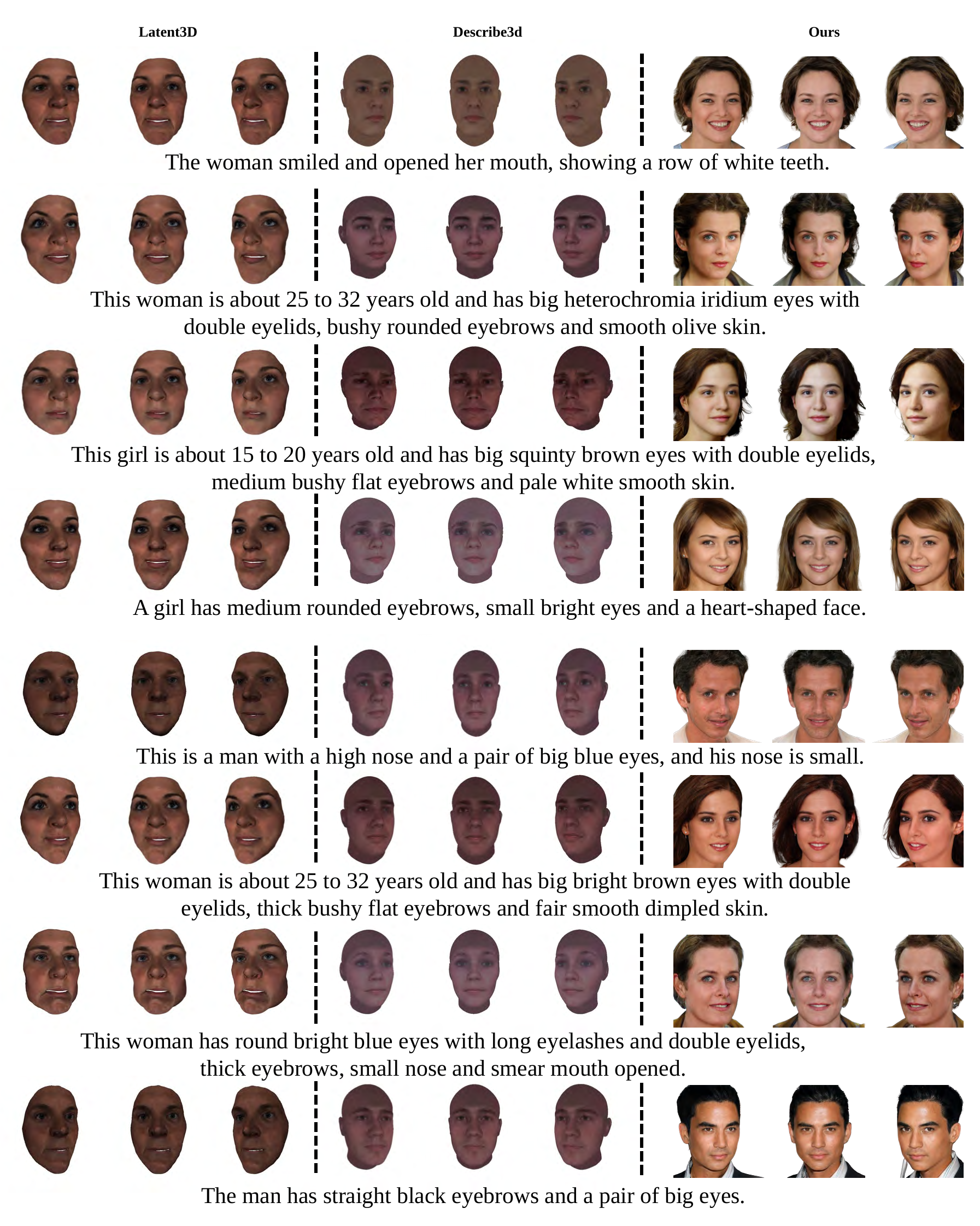}
    \caption{More comparison for 3D face generation. $E^3$-FaceNet excels in generating fine-grained facial attributes while maintaining multi-view consistency and preserving facial identity.}
    \label{fig:3D-gen}
\end{figure*}

\begin{figure*}[!t]
    \centering
    \includegraphics[width=1.0\linewidth]{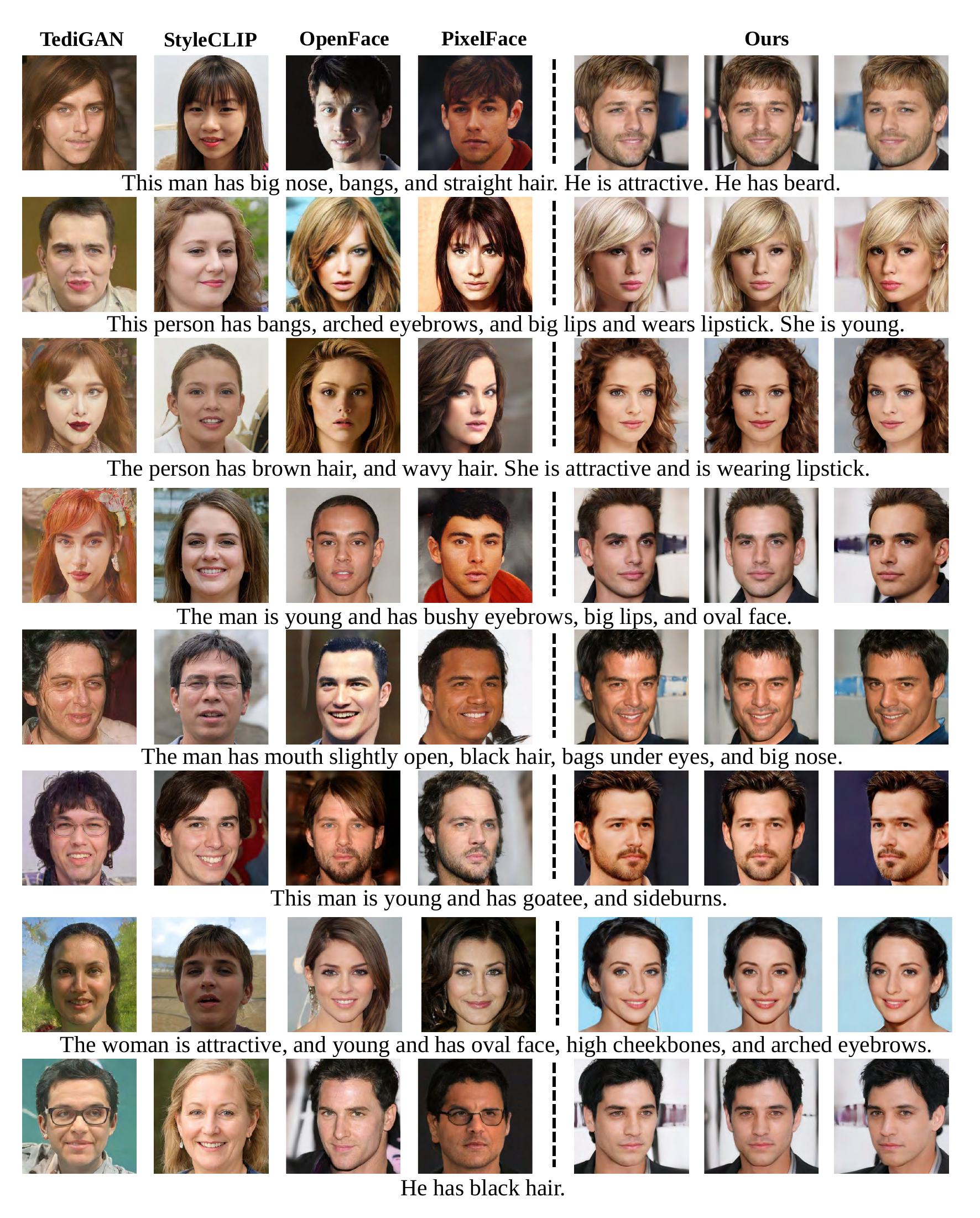}
    \caption{More comparisons with T2D Face methods. Our method showcases remarkable proficiency in maintaining high-quality visual output and ensuring semantic consistency across different views. }
    \label{fig:2D-gen}
\end{figure*}

We also present additional generation comparisons with both 3D and 2D face generation methods mentioned in the main paper in \cref{fig:3D-gen} and \cref{fig:2D-gen}. Our method excels in synthesizing high-quality 3D faces that align effectively with the given prompts and maintain consistency across various viewing angles. Additionally, our $E^3$-FaceNet method is capable of generating 3D faces with multiple facial assets, enhancing the diversity and realism of the generated results.

\begin{figure*}[!t]
    \centering
    \includegraphics[width=1.0\linewidth]{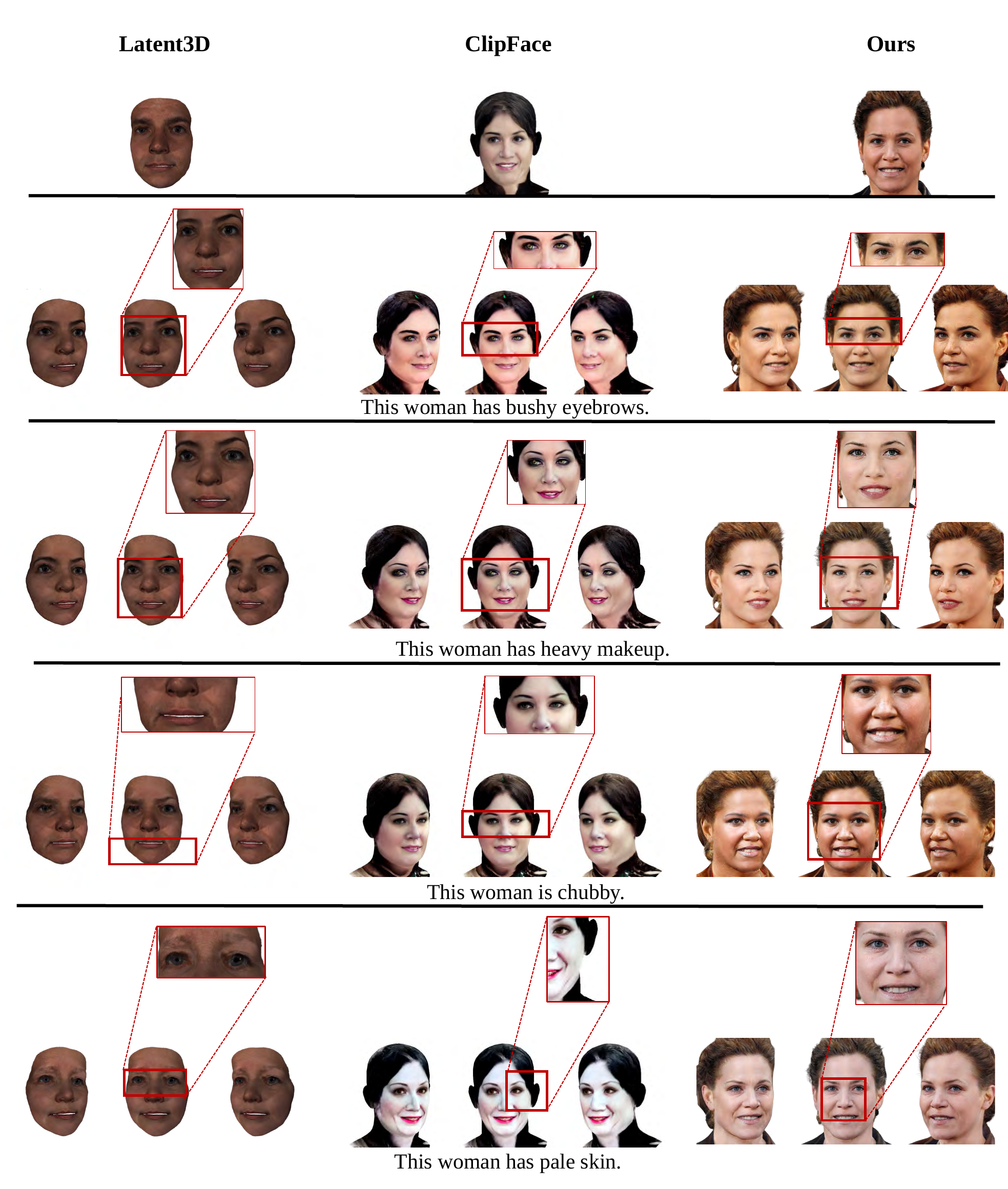}
    \caption{More comparisons for 3D face manipulation. $E^3$-FaceNet can faithfully edit the facial attributes described in the editing instructions while keeping other attributes unchanged. This ensures precise and targeted attribute manipulation in the generated 3D faces.}
    \label{fig:3D-edit}
\end{figure*}

More results for text-driven 3D Face manipulation in shown in \cref{fig:3D-edit}, in which we show some editing examples for female faces. This figure also demonstrates the limited edit ability of Latent3D, as neither \emph{``heavy makeup''} nor \emph{``pale skin''} is generated in its editions. As for ClipFace, the identity changes significantly and struggles to achieve multi-view consistency. As shown from editing results of \emph{``heavy makeup''}, ClipFace generates unnatural texture maps and the edited face appears distorted in the right view. In stark contrast, our $E^3$-FaceNet cannot only effectively respond to the text semantics but also maintain high image quality and multi-view consistency.

\begin{figure*}[!h]
    \centering
    \includegraphics[width=1.0\linewidth]{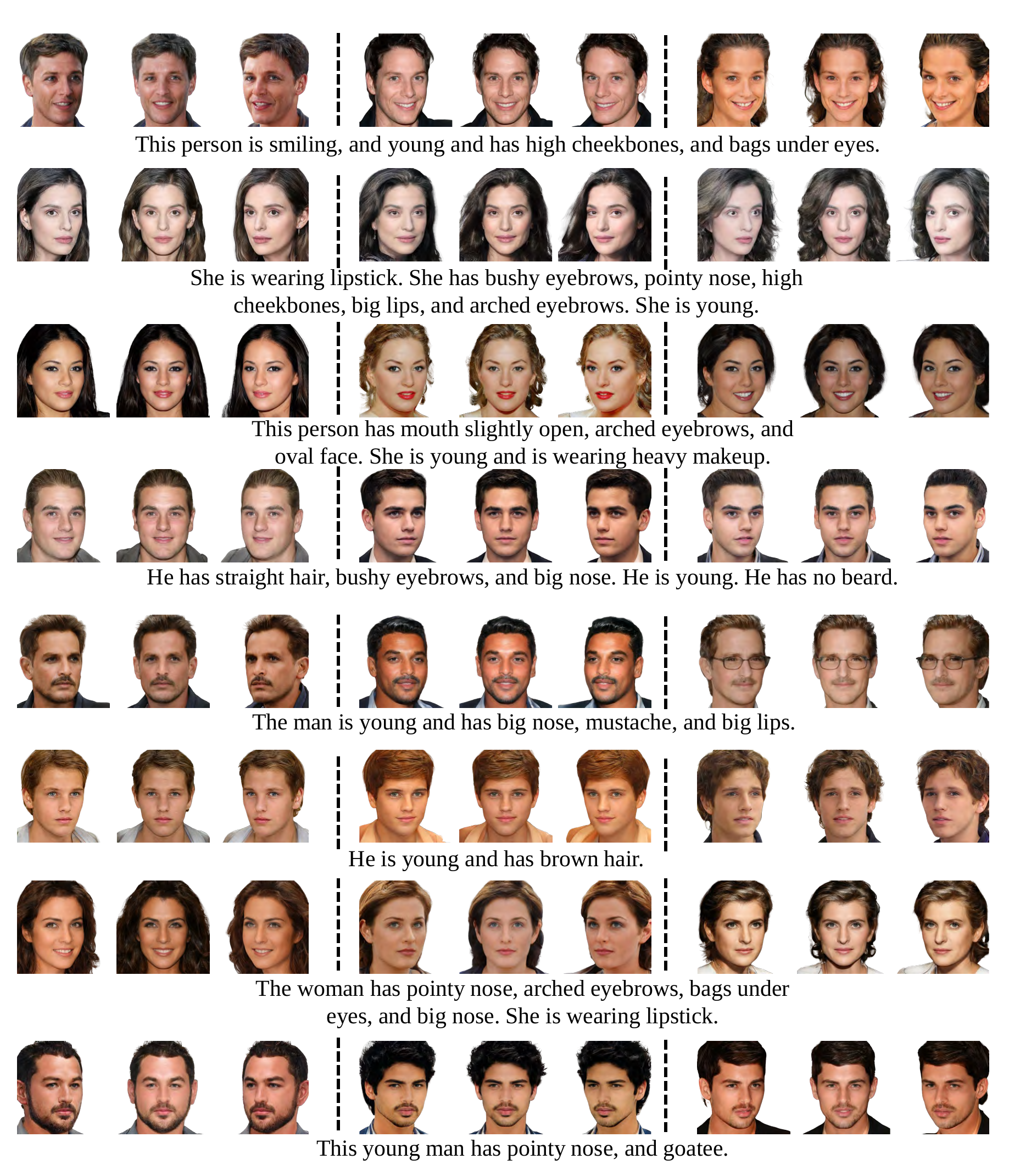}
    \caption{Diverse generation results of $E^3$-FaceNet. These results demonstrate that our proposed method can not only generate high-quality 3D faces that align well with the given prompts, but also achieve impressive diversity in the generated results.}
    \label{fig:3D-diverse}
\end{figure*}

\section{Additional Generation}
\label{sec:additional_results}
Meanwhile, our proposed $E^3$-FaceNet is capable of generating diverse 3D faces conditioned on the same input sentence, as demonstrated in \cref{fig:3D-diverse}. In each row, we sample different latent codes using different random seeds while inputting the same text prompts. Impressively and remarkably, these generated results showcase that $E^3$-FaceNet not only synthesizes high-quality and semantically aligned 3D faces but also possesses the ability for diverse 3D face generation.